\documentclass{bmvc2k}

\usepackage{graphicx}%
\usepackage{multirow}%
\usepackage{amsmath,amssymb,amsfonts}%
\usepackage{amsthm}%
\usepackage{mathrsfs}%
\usepackage[title]{appendix}%
\usepackage{xcolor}%
\usepackage{textcomp}%
\usepackage{manyfoot}%
\usepackage{booktabs}%
\usepackage{algorithm}%
\usepackage{algorithmicx}%
\usepackage{algpseudocode}%
\usepackage{listings}%

\usepackage{makecell}
\usepackage{array}
\usepackage{tabularx}
\usepackage{booktabs}

\usepackage{amsmath}
\usepackage{amssymb}
\usepackage{bbm}

\newcommand{\inlinehead}[1]{\noindent\textbf{#1}\quad}

\title{CountVideoBench: A Benchmark to Evaluate Prediction Biases on the Object Counting task by Video-Language Models}

\addauthor{Marc Chapus}{marc.chapus@liris.cnrs.fr}{1}
\addauthor{Carlos Crispim-Junior}{carlos.crispim-junior@liris.cnrs.fr}{1}
\addauthor{Véronique Eglin}{veronique.eglin@liris.cnrs.fr}{1}
\addauthor{Atilla Baskurt}{atilla.baskurt@liris.cnrs.fr}{1}

\addinstitution{
INSA Lyon,\\ Université Lumière Lyon 2,\\ CNRS, \\ Université Claude Bernard Lyon 1, \\
LIRIS, UMR5205, \\69621 Villeurbanne, France\\
}

\runninghead{Chapus, Crispim-Junior, Eglin, Baskurt}{CountVideoBench}

\begin{document}

\maketitle

\begin{abstract}
Object counting in videos remains a poorly characterized capability of multimodal large language models.
Counting questions are included in many existing VideoQA benchmarks, but the ground-truth count distributions in these benchmarks are heavily skewed toward small values.
Across the nine benchmarks we survey, the median ground-truth count is at most five, and the three most frequent values together account for between 45\% and 93\% of the videos.
The resulting mean accuracy is then dominated by the most frequent counts; a value-dependent bias at a rare count contributes negligibly to the average and remains invisible.
We introduce CountVideoBench, a benchmark of 2,600 synthetic videos covering 26 cardinalities in the interval [5, 50] (100 videos each), grouped into six clusters of consecutive counts that span the range while enabling near-neighbor comparisons.
Each video contains a single class of static objects under a controlled
camera motion, removing object motion and reducing recognition as confounds.
We evaluate sixteen video-language models from nine architecture families in a zero-shot setting.
No model exceeds $27.6\%$ Exact Match.
The dependence of accuracy on cardinality varies across models, ranging from monotonic decline to a non-monotonic peak-and-collapse pattern, which rules out increasing difficulty with cardinality as the dominant explanation of counting failure.
Every model further concentrates its predictions on its own set of preferred integers, with accuracy peaks coinciding with these values and collapsing at adjacent counts.
These findings highlight the need for balanced, diagnostic counting benchmarks to reliably assess this fundamental visual capability.
\end{abstract}

\section{Introduction}
\label{sec:intro}

Multimodal large language models now answer a wide range of open-ended questions over images and videos.
Counting is one such task, and an informative one to isolate: it is unambiguous, has exact ground truth, and exercises detection, instance separation, and temporal aggregation jointly.
It is also a documented weakness of the underlying vision encoders.
Image-text pretrained models such as CLIP~\cite{clip} encode cardinality poorly~\cite{paiss2023teaching, zhang2024can}, and the limitation persists in image-based vision-language models that build on these encoders~\cite{qharabagh2024lvlm}.
Video-language models inherit this limitation in turn, and add viewpoint change, occlusion, and temporal integration to a problem already unresolved on still images.

Counting in video has been studied directly~\cite{AminiNaieni25, lu2025avreasoner}, and appears in broader Video-QA benchmarks~\cite{li2024mvbench, liu2026vcbench} as one question type among many.
In both cases the count distributions concentrate most of their probability mass on a handful of small values, with long tails of rare large counts.
Accuracy averaged over such a distribution is dominated by the most frequent counts rather than reflecting counting ability across the intended range; any value-dependent bias at a rare count contributes negligibly to the mean and remains invisible.
Existing benchmarks are therefore not diagnostic of value-dependent counting behaviour.

We introduce CountVideoBench\footnote{Dataset and code: \url{https://countvideobench.github.io}. The videos are released under CC~BY~4.0; the generation code is released separately and does not redistribute third-party assets.}, a benchmark of $2{,}600$ synthetic videos designed to support such a diagnosis.
Videos are distributed uniformly over $26$ cardinalities in $[5, 50]$, with $100$ videos per cardinality, so that every cardinality contributes equally to any aggregate metric.
Each video contains a single class of static objects under a controlled camera motion, isolating counting from object recognition and object dynamics while preserving viewpoint variation.
The cardinalities form six clusters of consecutive values around the anchors $\{5, 10, 20, 30, 40, 50\}$, with adjacent counts inside each cluster to enable near-neighbor comparisons.
We evaluate sixteen video-language models from nine architecture families in a zero-shot setting, using a single prompt and the same generation length limit for all models.

\begin{figure}[!t]
    \centering

    \makebox[\textwidth][c]{%
        \includegraphics[width=0.32\textwidth]{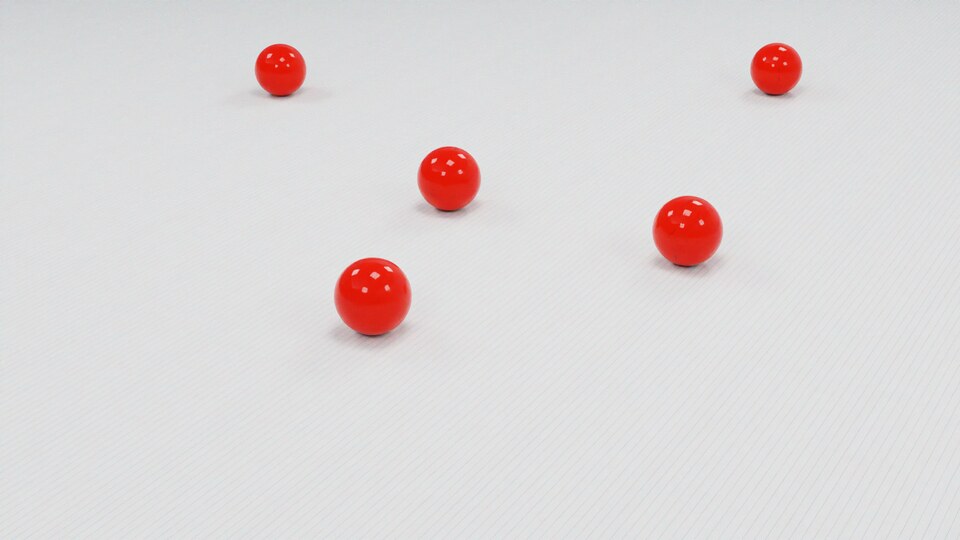}\hfill
        \includegraphics[width=0.32\textwidth]{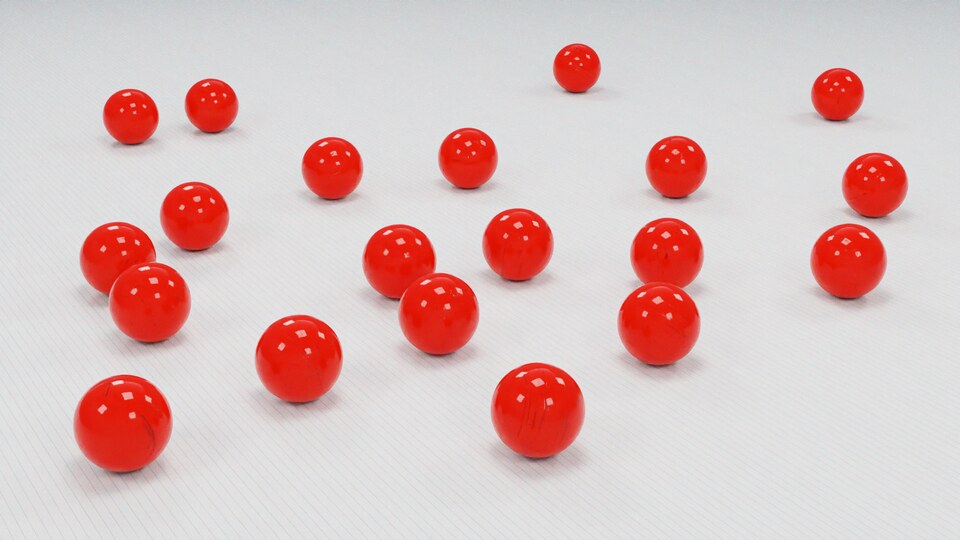}\hfill
        \includegraphics[width=0.32\textwidth]{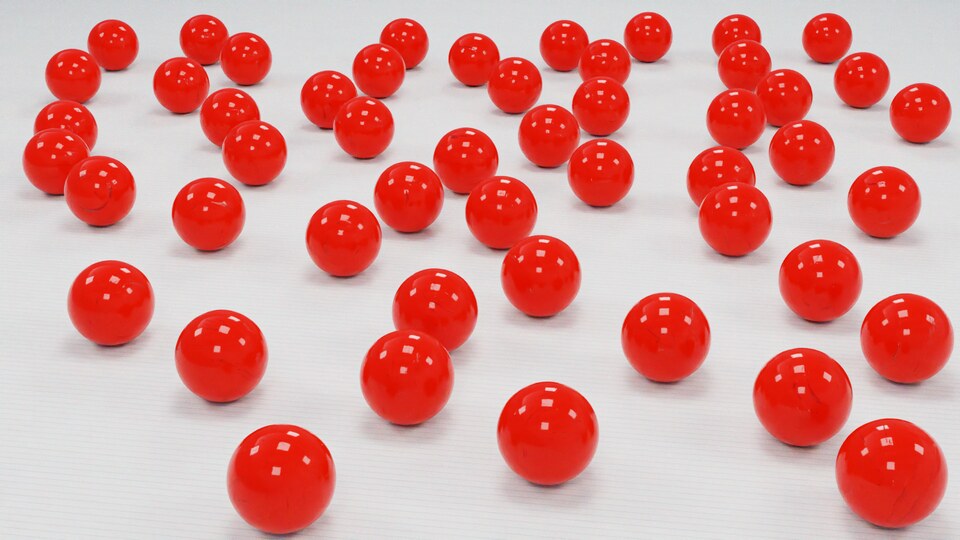}%
    }%

    \vspace{0.4em}

    \makebox[\textwidth][c]{%
        \includegraphics[width=0.32\textwidth]{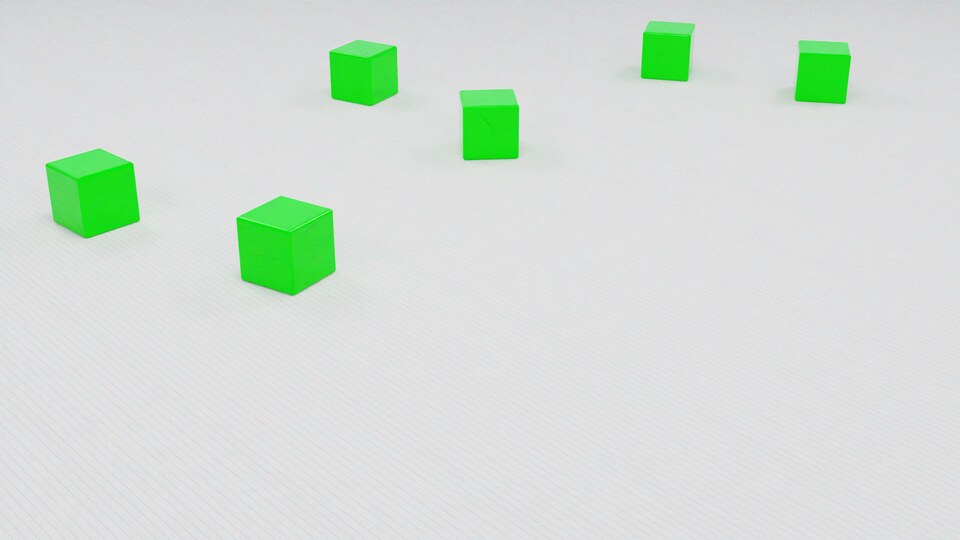}\hfill
        \includegraphics[width=0.32\textwidth]{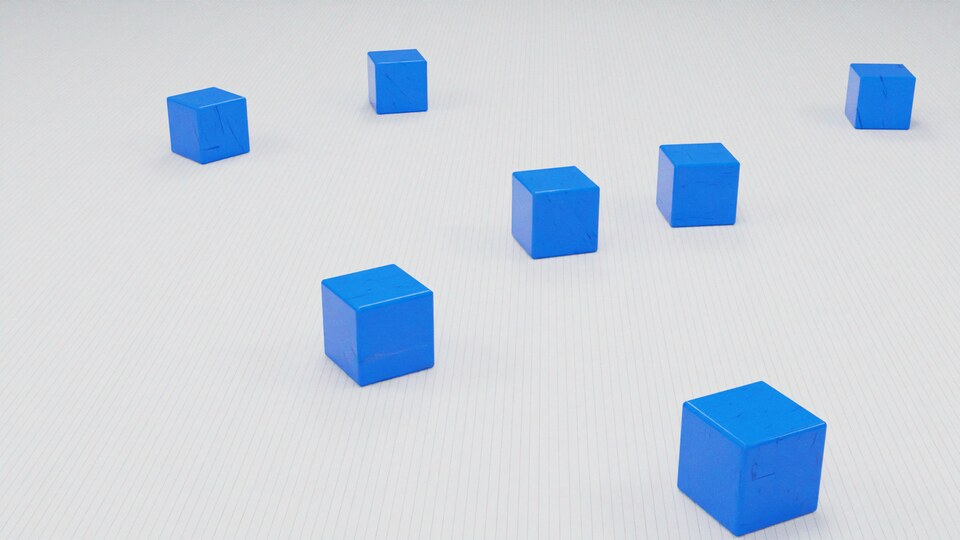}\hfill
        \includegraphics[width=0.32\textwidth]{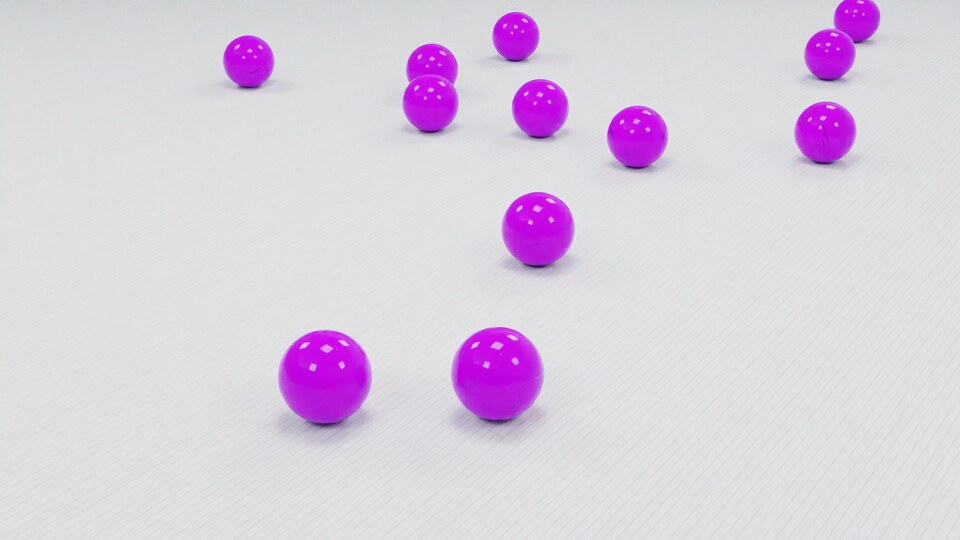}%
    }%

    \caption{Example frames from CountVideoBench illustrating coverage across cardinality (top) and object categories (bottom). Camera motion and isolated object classes isolate counting from recognition and object motion while preserving viewpoint variation.}
    \label{fig:example_frames}
\end{figure}

Three findings emerge.
First, counting remains an unsolved problem even in this minimalist setting: no model exceeds $27.6\%$ Exact Match, and most stay well below this ceiling.
Second, the dependence of accuracy on cardinality varies across models: some decline monotonically, others show isolated peaks at specific cardinalities with near-zero accuracy at adjacent counts.
This rules out cardinality-induced difficulty as the dominant cause of counting failure; the variation is measurable here only because every cardinality contributes the same number of videos.
Third, every model concentrates its predictions on its own set of preferred integers, with accuracy peaks coinciding with these values and collapsing at adjacent counts.
In most models, predictions are confined to a small set of preferred values; in those with the broadest output distributions, including the two flagship proprietary ones, the concentration appears instead as local irregularities.

\section{Related Work}\label{sec:related}

\subsection{Video Question Answering with Multimodal LLM}
\label{sec:rw_models}

Video question answering was first addressed with task-specific architectures jointly training a video encoder, a question encoder, and a fusion module on supervised Video-QA data~\cite{le2020hierarchical,yang2021justask,yang2022frozenbilm}.
The current generation is built on instruction-tuned multimodal large language models, and our fourteen open-source models span three design directions.
NVILA~\cite{liu2025nvila} and AdaReTaKe~\cite{wang2025AdaReTaKe} target long-video processing through visual-token compression, learned and training-free respectively.
Qwen2.5-VL~\cite{Qwen2.5-VL}, Qwen3-VL~\cite{Qwen3-VL} and InternVL2.5~\cite{chen2024expanding} extend image vision-language models to native video, through unified dynamic-resolution encoders or progressive transfer to a larger language model.
VideoLLaMA2~\cite{damonlpsg2024videollama2} and LLaVA-Video~\cite{zhang2024llava} treat video as the primary modality, through a dedicated spatio-temporal connector and video-centric instruction tuning respectively.
Two proprietary models, GPT-5.4~\cite{openai2025gpt5card} and Gemini-3.1-pro~\cite{geminiteam2023gemini}, complete the evaluation; their architectures are undisclosed but widely understood to exceed our largest open-source variants in parameter count.

\subsection{Counting in Vision-Language Models}
\label{sec:rw_counting}

Counting is a documented weakness of vision-language models built on image-text pretrained vision encoders. Paiss et al.~\cite{paiss2023teaching} show that CLIP \cite{clip} fails on the simple task of associating an image with the prompt \emph{``a photo of $N$ objects''} for small $N$, and attribute the failure to a pretraining objective that supplies essentially no supervision for cardinality.
Zhang et al.~\cite{zhang2024can} characterize a systematic quantity bias in the CLIP representation, where predicted counts concentrate on a small set of frequent values across object categories.
Qharabagh et al.~\cite{qharabagh2024lvlm} extend the analysis to large vision-language models and find that the encoder's counting limitation propagates through the language-model interface, with these models failing on moderate cardinalities even on still images.

Video-language models inherit this constraint by construction: their vision encoders (CLIP, SigLIP~\cite{siglip}, InternViT~\cite{chen2024expanding}, and other image-text pretrained ViTs) were trained on large-scale image-text pairs that supply essentially no supervision for cardinality.

Temporal aggregation therefore operates on features that already encode cardinality poorly at the frame level, and adds viewpoint changes, partial occlusion, and temporal integration to a problem that is unresolved on still images.
Closest to our setting, VLMCountBench~\cite{guo2025vlmcountbench} isolates counting on still images using minimal geometric shapes under strict variable control, and reports reliable counting when a single shape type is present.
We transpose this minimalist philosophy to video and vary cardinality rather than compositional complexity, and find that the single-class regime does not remain reliable over a range extending to fifty.
The benchmarks through which counting in video is currently evaluated are reviewed next.

\subsection{Counting Benchmarks in Video}
\label{sec:rw_benchmarks}

Counting questions appear in a number of video benchmarks. Four of them are recurrent reference points in the recent counting literature.

MVBench~\cite{li2024mvbench} is a comprehensive video-understanding benchmark spanning twenty perception and reasoning tasks; counting appears as multiple-choice sub-tasks with object and action variants over a small set of candidate answers.

VideoCount~\cite{AminiNaieni25} casts counting in video as an open-world task: a system receives a text description or a visual exemplar of the target class and must enumerate the unique instances appearing in the clip.
The benchmark unifies subsets of the tracking datasets TAO~\cite{dave2020tao} and MOT20~\cite{dendorfer2020mot20} with scientific footage of penguins and metal-alloy crystallization.
It does report results by count range, but only across these sub-benchmarks (TAO: $1$--$10$; MOT20: $80$--$1203$; Science: $10$--$154$): strata separated by orders of magnitude cannot reveal value-dependent behavior between consecutive cardinalities.

CG-AV-Counting~\cite{lu2025avreasoner} targets long-form videos with clue-grounded multimodal queries; counting targets cover events, objects, and attributes, and queries combine audio-only, visual-only, and joint audio-visual formats.
VCBench~\cite{liu2026vcbench} introduces streaming counting as a probe for spatio-temporal state maintenance, decomposing the task into object and event variants (currently visible instances, cumulative unique identities, instantaneous actions, and complete activity cycles) and querying the model at multiple points along the timeline.

Counting also appears as one question type among many in NExT-QA~\cite{xiao2021next}, TGIF-QA~\cite{jang-IJCV-2019}, and Video-MME~\cite{fu2025video}, and as a derived sub-task in the synthetic-video benchmarks CLEVRER~\cite{CLEVRER2020ICLR} and VLM4D~\cite{zhou2025vlm4d}.

Despite this diversity, none of these benchmarks is specifically designed for diagnostic evaluation of object-instance counting across a broad and balanced range of cardinalities.
We quantify this gap in the next section, where we measure the imbalance in the count distribution of existing benchmarks and motivate the design of CountVideoBench against it.

\section{Empirical Analysis of Benchmark Biases} \label{sec:problem_analysis}

Accuracy on a counting benchmark is a weighted average over the cardinalities present in the data.
When the count distribution is concentrated on a small number of values, the resulting score reflects behavior on those values rather than counting ability across the intended range. A model that counts well on the dominant cardinalities and poorly elsewhere can yield the same aggregate score as one with balanced behavior.
Counting failures in video-language models need not be uniform along the cardinality axis: a model may concentrate predictions on a small set of preferred values, or exhibit accuracy that varies non-monotonically with the true count.
Detecting such patterns therefore requires each cardinality of interest to receive enough videos for a stable per-value estimate.
This fails when most counts appear only once or twice and most videos sit on a handful of small values.

To quantify the imbalance, we report for each benchmark its \emph{Top-3 share}: the proportion of videos whose ground-truth count falls within the three most frequent cardinalities.
The statistic is independent of the count range and number of distinct values, and varies from $3/k$ on a strictly uniform benchmark with $k$ distinct counts to $100\%$ on one whose mass collapses onto three values.
Table~\ref{tab:dist_summary} reports the Top-3 share, the count range, the number of distinct counts, and the median with interquartile range for nine existing video benchmarks that include counting questions. 
Figure~\ref{fig:dist_existing} plots three representative ones; all nine plus ours appear in Appendix Fig.~\ref{fig:dist_existing_annexe}.
All three share a heavy concentration of probability mass on counts below ten, with Top-3 shares between 45.3\% (VCBench~\cite{liu2026vcbench}) and 93.1\% (VideoCount~\cite{AminiNaieni25}).
VideoCount, CG-AV-Counting~\cite{lu2025avreasoner}, and VCBench additionally exhibit long tails of rare large counts with negligible statistical weight.
MVBench~\cite{li2024mvbench} is capped at 9 and exhibits strong mass concentration even within this short range, with the single value $3$ accounting for $33\%$ of the benchmark.
Across all nine benchmarks in Table~\ref{tab:dist_summary}, the median ground-truth count is at most five and the Top-3 share never falls below $45\%$.

\begin{figure}[!t]
    \centering
    \includegraphics[width=\linewidth]{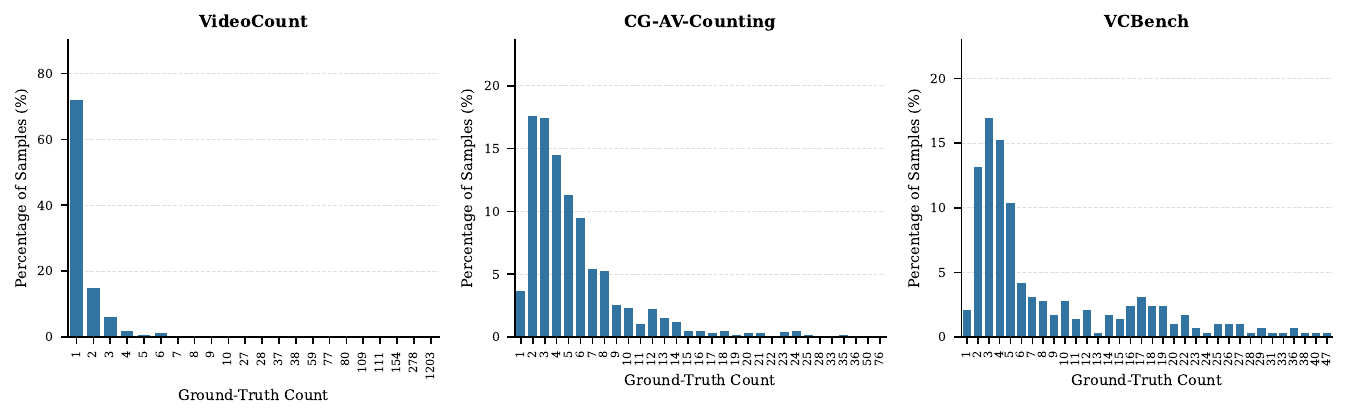}
    \caption{Ground-truth count distributions in three video benchmarks with counting questions: VideoCount, CG-AV-Counting, and VCBench. Probability mass concentrates on small counts, with long tails of rare large counts carrying negligible statistical weight.}
    \label{fig:dist_existing}
\end{figure}

\begin{table}[!t]
\centering
\caption{Distribution characteristics of the ground-truth count variable in video benchmarks with counting questions. CountVideoBench is reported in the last row.}
\label{tab:dist_summary}
\begin{tabular}{lcccc}
\toprule
Benchmark & Range & Distinct counts & Med [Q1--Q3] & Top-3 share \\
\midrule
MVBench~\cite{li2024mvbench}             & 0--9      & 10 & 3 [2--3]    & 66.5\% \\
NExT-QA~\cite{xiao2021next}              & 1--15     & 13 & 2 [2--4]    & 72.5\% \\
TGIF-QA~\cite{jang-IJCV-2019}              & 0--10     & 10 & 2 [0--4]    & 67.9\% \\
Video-MME~\cite{fu2025video}          & 0--600000 & 25 & 4 [2--6]    & 53.5\% \\
VideoCount~\cite{AminiNaieni25}       & 1--1203   & 22 & 1 [1--2]    & 93.1\% \\
CG-AV-Counting~\cite{lu2025avreasoner} & 1--76     & 31 & 4 [3--7]    & 49.6\% \\
VCBench~\cite{liu2026vcbench}             & 1--47     & 34 & 5 [3--12]   & 45.3\% \\
CLEVRER~\cite{CLEVRER2020ICLR}             & 0--5      &  6 & 1 [1--2]    & 75.7\% \\
VLM4D~\cite{zhou2025vlm4d}                 & 0--90     & 12 & 1 [0--2]    & 78.0\% \\
\midrule
CountVideoBench (Ours)             & 5--50     & 26 & 25 [11--39] & 11.5\% \\
\bottomrule
\end{tabular}
\end{table}

CountVideoBench is designed to address this constraint. Section~\ref{sec:dataset} describes its construction in detail; for the present analysis the relevant feature is that the benchmark distributes $2{,}600$ videos uniformly over $26$ cardinalities in the range $[5, 50]$, with $100$ videos per cardinality.
The resulting Top-3 share is $11.5\%$, matching the $3/26\approx 11.5\%$ lower bound for a uniform design over $26$ cardinalities, and substantially below every benchmark in Table~\ref{tab:dist_summary}.
Metrics computed on this distribution therefore weight every cardinality equally, and the value-dependent biases analysed in Section~\ref{sec:bias} remain visible rather than absorbed into a few dominant counts.

\section{CountVideoBench Dataset}
\label{sec:dataset}

\subsection{Design Principles}
\label{sec:design_principles}

CountVideoBench follows two design principles addressing the limitations of Section~\ref{sec:problem_analysis}.

The first principle is \emph{distributional balance}.
Each cardinality used for evaluation is represented by the same number of videos.
Aggregate metrics therefore weight every count equally.
The resulting Top-3 share (last row of Table~\ref{tab:dist_summary}) is approximately four times smaller than that of the most balanced existing benchmark in our comparison.

The second principle is \emph{isolation from recognition}.
Counting in natural video entangles detection, classification, instance separation, and temporal aggregation. To diagnose counting itself, we reduce recognition and motion as confounds:
each video contains a single class of simple geometric objects, and objects are static while the camera moves.
A model is therefore evaluated on enumerating instances of an already-identified class, and failures in this setting are unlikely to stem primarily from recognition or motion.

\subsection{Anchor-Neighborhood Structure}
\label{sec:anchor}

The cardinality sampling is shaped by two requirements.
The benchmark should span more than an order of magnitude in cardinality, and support bias detection through near-neighbor comparisons where the task and scene remain otherwise comparable.
A continuous uniform distribution over $[5, 50]$ would satisfy both requirements but would require $46 \times 100 = 4{,}600$ videos, beyond our rendering budget.
We adopt an anchor-neighborhood structure that preserves the range while reducing the rendering cost, and concentrates sampling where neighbour comparisons are informative.

Six anchor cardinalities are selected, $\mathcal{A} = \{5, 10, 20, 30, 40, 50\}$.
For each anchor $n \in \mathcal{A}$, the five consecutive cardinalities $\{n-2, n-1, n, n+1, n+2\}$ are included; the union is bounded to $[5, 50]$.
The four interior anchors ($10$, $20$, $30$, $40$) thus contribute five cardinalities each; the two edge anchors ($5$ and $50$) lose two values to truncation and contribute three each.
The benchmark covers $26$ distinct cardinalities, one hundred videos each, yielding $2{,}600$ videos in total. We refer to each anchor and its neighbours as a \emph{cluster}.

Within each cluster, the five neighboring cardinalities share the same task and differ only by small amounts.
Comparing accuracy across a cluster therefore probes model behavior under near-identical inputs, and reveals whether predictions track the ground truth or concentrate at specific values: at the anchor, elsewhere in the cluster, or outside.
Dense sampling within each cluster provides the local resolution to read these patterns.

\subsection{Generation Pipeline and Composition}
\label{sec:pipeline}

Videos are generated procedurally in Blender, each scene fully determined by a single random seed. Each video lasts five seconds at $30$ frames per second ($150$ frames) and is rendered at $1920 \times 1080$. Figure~\ref{fig:example_frames} shows example frames at three cardinalities.

\inlinehead{Object classes.} The benchmark uses ten object classes, crossing two shapes, cube and sphere, with five colors: red, blue, green, yellow, and purple (Figure~\ref{fig:example_frames}).
Each video contains a single class, so the counting question admits a unique correct answer.
The colors are mutually distinguishable against the white background and avoid chromatic adjacencies (e.g.\ orange/yellow, cyan/blue) that could conflate counting with color identification. Both shapes and colors remain identifiable at small angular sizes.

\inlinehead{Scene and camera.} The background is uniform white overlaid with a regular line pattern, and object surfaces carry a low-amplitude procedural noise that produces faint per-instance texture variations. These choices address a difficulty at high density: perfectly identical, densely packed objects visually merge and become hard to enumerate even to a human observer. The line pattern provides a spatial reference and the surface noise breaks visual symmetry between adjacent instances, without altering color or shape identification.
Lighting is uniform and identical across all videos.
Objects are placed at random positions within a bounded region of the ground plane; a minimum-distance constraint combined with Blender's collision system prevents overlap (Appendix~\ref{app:pipeline}).
The camera rotates in place, with orientation changes drawn within bounded angular ranges.
The placement bounds and rotation range are chosen so that every object is visible in at least some frames, keeping the count attainable, while no single frame is guaranteed to show all $N$ at once: $83.0\%$ of videos contain at least one object leaving the field of view, and on average $3.29$ objects per video fall outside the frame at some point during the clip.
The task therefore requires integrating partial views over time to avoid omissions and double counts.

\inlinehead{Composition.} Within each cardinality, the ten object classes are equally represented, with exactly ten videos per (shape, color) class. The random seed governing object positions and the camera trajectory is drawn independently for each video.
Each video is paired with a single question--answer pair, generated automatically from the scene metadata, which ensures exact ground-truth availability without any human annotation step.

\section{Experiments}\label{sec:protocol}

\subsection{Models and Inference}
\label{sec:models}

On CountVideoBench, we evaluate sixteen video-language models drawn from nine architecture families.
Seven of these families are open-source: AdaReTaKe~\cite{wang2025AdaReTaKe}, LLaVA-Video~\cite{zhang2024llava}, NVILA~\cite{liu2025nvila}, Qwen2.5-VL~\cite{Qwen2.5-VL}, Qwen3-VL~\cite{Qwen3-VL}, VideoLLaMA2~\cite{damonlpsg2024videollama2}, and InternVL2.5~\cite{chen2024expanding}.
Each family contributes a lightweight variant ($7$--$8$B parameters) and a full-scale variant ($15$--$78$B).
This pairing enables comparing counting behavior across scale within a fixed architecture.
Two proprietary models, GPT-5.4~\cite{openai2025gpt5card} and Gemini-3.1-pro~\cite{geminiteam2023gemini}, are included only in their flagship configuration, as high-capacity reference points; smaller variants lack the public documentation needed for a controlled scale comparison.

All models are evaluated zero-shot on ten frames sampled uniformly from each clip, with no in-context examples and no fine-tuning on counting data.
A single prompt is used: \emph{``How many [color] [shape]s are in this video?''}, with \texttt{[color]} and \texttt{[shape]} filled from the scene metadata.
The decoding budget is set to \texttt{max\_tokens}~$=1024$ for all models.
Responses are typically short natural-language sentences containing the predicted count, occasionally with a brief justification.

Model outputs are free-form text and do not always consist of a single integer.
We extract the predicted count from each response with a lightweight model parser (Qwen2.5-7B~\cite{qwen2}), returning the integer when one can be unambiguously identified and \texttt{NaN} otherwise. The full extracted CSV is then verified manually.
Predictions tagged \texttt{NaN} represent non-responses (refusals, qualitative answers such as \emph{``many''}, generations that do not commit to a numerical value, or ambiguous outputs containing multiple incompatible integers) and are treated separately in the metrics defined below.

\subsection{Evaluation Metrics}
\label{sec:metrics}

Three primary metrics target complementary aspects of counting behavior.
Exact Match measures strict numerical correctness.
Tolerance Accuracy captures approximate correctness within a value-dependent margin.
Coverage measures whether a numerical answer is produced, disentangling wrong answers from abstention as causes of low Exact Match.

Let $\hat{y}_i$ denote the predicted count for video $i$ and $y_i$ its ground-truth count, with $i \in \{1,\dots,N\}$ and $N=2{,}600$.
Exact Match is the proportion of videos for which the prediction is numerically equal to the ground truth, non-responses ($\hat{y}_i = \mathtt{NaN}$) are counted as incorrect.
\begin{equation}
\mathrm{EM} = \frac{1}{N}\sum_{i=1}^{N}\mathbf{1}\!\left[\hat{y}_i = y_i\right].
\label{eq:em}
\end{equation}

Tolerance Accuracy at $10\%$ uses a value-dependent margin $\tau(y) = \lfloor 0.1\,y \rfloor$:
\begin{equation}
\mathrm{Acc}@10\% = \frac{1}{N}\sum_{i=1}^{N}\mathbf{1}\!\left[\,|\hat{y}_i - y_i| \le \tau(y_i)\right].
\label{eq:acc10}
\end{equation}
The floor operation makes $\tau(y) = 0$ for $y \in \{5,\dots,9\}$, so
$\mathrm{Acc}@10\%$ coincides with $\mathrm{EM}$ on this range; tolerance
becomes effective from $y = 10$ onwards.

Coverage is the proportion of videos for which a numerical answer is produced,
\begin{equation}
\mathrm{Cov} = \frac{1}{N}\sum_{i=1}^{N}\mathbf{1}\!\left[\hat{y}_i \neq \mathtt{NaN}\right].
\label{eq:cov}
\end{equation}
A high Coverage with low accuracy reflects systematic but incorrect predictions, whereas a low Coverage indicates non-numeric, ambiguous, or abstained answers.

We complement these with the \emph{support size} (number of distinct cardinalities a model uses) and the prediction \emph{perplexity} (how concentrated those uses are). Both support the bias analysis in Section~\ref{sec:bias}.
Let $p_k$ denote the empirical frequency of output $k$, non-responses forming one extra category not counted in $|\mathcal{S}|$.
The support size is the number of cardinalities receiving non-zero mass,
\begin{equation}
|\mathcal{S}| = \bigl|\{k : p_k > 0\}\bigr|,
\label{eq:support}
\end{equation}
and gives a coarse measure of output diversity that is independent of correctness.
The prediction perplexity is the exponentiated Shannon entropy of the prediction distribution,
\begin{equation}
\mathrm{PP} = 2^{H(p)}, \qquad H(p) = -\sum_{k} p_k \log_2 p_k,
\label{eq:pp}
\end{equation}
and corresponds to the effective number of equally-frequent counts in the prediction distribution.
A model that always answers a single value has $\mathrm{PP} = 1$; a uniform distribution over the $26$ benchmark cardinalities yields $\mathrm{PP} = 26$.
The two descriptors are complementary: a model can have a large support while concentrating most of its predictions on a small subset of it, in which case $\mathrm{PP}$ remains far below $|\mathcal{S}|$.

$|\mathcal{S}|$ and $\mathrm{PP}$ summarize the prediction distribution globally; two further quantities measure its local irregularities.
For each integer value $k$, let $N_k$ denote the number of times $k$ is predicted, and let $L_k = \max(N_{k-1}, N_{k-2})$ and $R_k = \max(N_{k+1}, N_{k+2})$ denote the largest prediction count on each side. The \emph{local excess} at $k$ is $N_k - \max(L_k, R_k)$, positive when $k$ stands above all four neighbours. The \emph{local deficit} at $k$ is $\min(L_k, R_k) - N_k$, positive only when $k$ sits in a valley between two flanking peaks. We report the largest excess and deficit observed across each model's support.

\section{Results and Discussion}
\label{sec:results}

\subsection{Overall Performance}
\label{sec:overall}

Table~\ref{tab:results} reports overall performance and output diversity on CountVideoBench.

\inlinehead{Exact Match.} Three models cluster at the top of the Exact Match (EM) column within two points of each other: Gemini-3.1-pro at $27.6\%$, LLaVA-Video-7B at $26.8\%$, and GPT-5.4 at $25.7\%$ (all results use a single prompt formulation fixed in advance, Section~\ref{sec:protocol}).
No other model exceeds $23\%$; five of sixteen stay below $10\%$.
Even on a benchmark with a single object class, static instances, and exact ground truth, agreement with the true cardinality remains uncommon.
One model behaves degenerately: VideoLLaMA2-7B answers only $1.2\%$ of videos, collapsing its EM to $0.1\%$.
Inspection shows that the model most often returns a refusal or a sentence without a count rather than a wrong number, indicating an instruction-following failure rather than a counting failure.

\inlinehead{Human reference.} On a subset of $104$ videos, twenty participants viewing the same ten sampled frames as the models reach $90.8\%$ mean EM against $26.9\%$ for the best models, with a gradual decline across cardinality and no peak-and-collapse structure.
The scenes are countable; the low model scores reflect limitations of current approaches (Appendix~\ref{app:human}).

\inlinehead{Within-family scaling is inconsistent.} The two proprietary models, Gemini-3.1-pro and GPT-5.4, place first and third on EM, and both are widely understood to operate well beyond the $72$ to $78$B range of our largest
open-source variants, which would invite a scale-based reading of their lead.
The within-family numbers caution against it. Across the seven open-source families, four place the larger variant ahead by $2$ to $6$ points, two of them in regimes where both variants stay below $6\%$, two differ by less than two points, and one (LLaVA-Video) places the $7$B variant ahead by $12.7$. The largest within-family gap therefore goes in the inverse-scaling direction, and the most accurate open-source model on EM is the $7$B variant of its family.
Scale is presumably part of the proprietary lead, but it is not alone a reliable predictor of counting accuracy here; training data composition and post-training alignment plausibly contribute, and our benchmark does not isolate these factors.

\inlinehead{Magnitude metrics do not capture prediction structure.} MAE and RMSE, computed over answered videos, summarize how far predictions fall from the truth but not how they are distributed: LLaVA-Video-7B and NVILA-15B share an RMSE of $6.1$ and differ by $0.2$ in MAE, yet differ by a factor of two in EM ($26.8\%$ against $12.5\%$) and in concentration ($\mathrm{PP}$ $8.0$ against $14.3$).
They complement EM without exposing the preferred-value concentration and adjacent-cardinality collapses of Section~\ref{sec:bias}.

\inlinehead{Errors are typically near misses.}
The gap between EM and Acc@10\% is substantial for most models, and the ranking on the relaxed metric differs from the strict one.
GPT-5.4 stands out at $70.5\%$, nearly $45$ points above its own EM and more than $14$ points ahead of the next model.
Eleven of the sixteen models gain more than $13$ points under tolerance.
In applications tolerating a small relative error, such as crowd-level summaries, several models reach usable accuracy.
Acc@10\% does not rescue every model, however: four stay below $10\%$ even with the relaxed criterion (Table~\ref{tab:results}). For VideoLLaMA2-7B this reflects its near-zero coverage; for the three others, errors are large in relative magnitude.

\inlinehead{Two response policies behind Exact Match.} Coverage varies more widely across models than either accuracy measure, and reveals two distinct response policies.
Eight models answer at least $97\%$ of videos and effectively never abstain; abstention rises steadily across the other eight, from under $9\%$ to half the benchmark for InternVL2.5-78B.
Two regimes therefore coexist behind the Exact Match column. Systematic answering, with predictions that are often wrong but always present, is exemplified by NVILA-8B at $100\%$ Coverage and $6.6\%$ Exact Match.
Selective abstention, with comparatively higher per-answered accuracy, is exemplified by LLaVA-Video-72B at $69.1\%$ Coverage and $14.1\%$ Exact Match.
Comparing models on Exact Match alone conflates these regimes, and Section~\ref{sec:cardinality} shows that abstention is structured along the cardinality axis rather than uniform.

\begin{table}[t]
\centering
\caption{Results on CountVideoBench. EM, Acc@10\% and Coverage in percent; MAE and RMSE in object-count units (lower is better). Bold marks the best value per column; for MAE/RMSE it ignores $\dagger$, whose Coverage is $1.2\%$. Oracle: exact prediction everywhere.}
\label{tab:results}
\begin{tabular}{lccccccc}
\toprule
Model & EM & Acc@10\% & MAE & RMSE & Coverage & $|\mathcal{S}|$ & $\mathrm{PP}$ \\
\midrule
AdaReTaKe-7B      & 15.1 & 34.4 &  4.5 &   6.8 &  91.1 & 28 & 11.3 \\
AdaReTaKe-72B     & 16.7 & 39.1 &  4.2 &   6.2 &  97.5 & 27 & 12.3 \\
InternVL2.5-8B    & 14.0 & 38.2 & 15.8 &  28.5 &  97.9 & 16 &  9.4 \\
InternVL2.5-78B   & 15.8 & 23.4 &  4.5 &   8.5 &  49.7 & 17 &  5.9 \\
LLaVA-Video-7B    & 26.8 & 56.3 &  3.7 &   6.1 &  98.9 & 16 &  8.0 \\
LLaVA-Video-72B   & 14.1 & 27.8 &  2.5 &   3.4 &  69.1 & 21 & 10.5 \\
NVILA-8B          &  6.6 & 24.5 &  5.9 &   9.1 & 100.0 & 26 & 12.8 \\
NVILA-15B         & 12.5 & 45.4 &  3.5 &   6.1 & 100.0 & 27 & 14.3 \\
Qwen2.5-VL-7B     &  1.9 &  6.1 &  8.6 &  11.8 &  73.7 & 32 & 10.9 \\
Qwen2.5-VL-72B    &  3.9 &  8.9 &  7.4 &  10.0 &  82.2 & 31 & 17.6 \\
Qwen3-VL-8B       & 18.7 & 47.7 &  3.4 &   5.5 & 100.0 & 31 & 17.7 \\
Qwen3-VL-32B      & 22.1 & 39.4 &  4.5 &   9.3 &  75.1 & 83 & 27.8 \\
VideoLLaMA2-7B$^\dagger$    &  0.1 &  0.1 &  1.7 &   1.9 &   1.2 &  5 &  1.1 \\
VideoLLaMA2-72B   &  5.0 &  7.6 & 11.9 &  17.8 &  71.1 & 25 &  8.9 \\
\midrule
GPT-5.4           & 25.7 & \textbf{70.5} & \textbf{1.7} & \textbf{2.4} & 100.0 & 48 & 30.6 \\
Gemini-3.1-pro    & \textbf{27.6} & 55.3 &  3.1 &   5.2 &  97.7 & 55 & 38.9 \\
\midrule
Oracle            & 100.0 & 100.0 &  0.0 &   0.0 & 100.0 & 26 & 26.0 \\
\bottomrule
\end{tabular}
\end{table}

\subsection{Behavior Across the Cardinality Range}
\label{sec:cardinality}

\begin{figure}[!t]
    \centering
    \includegraphics[width=1.\textwidth]{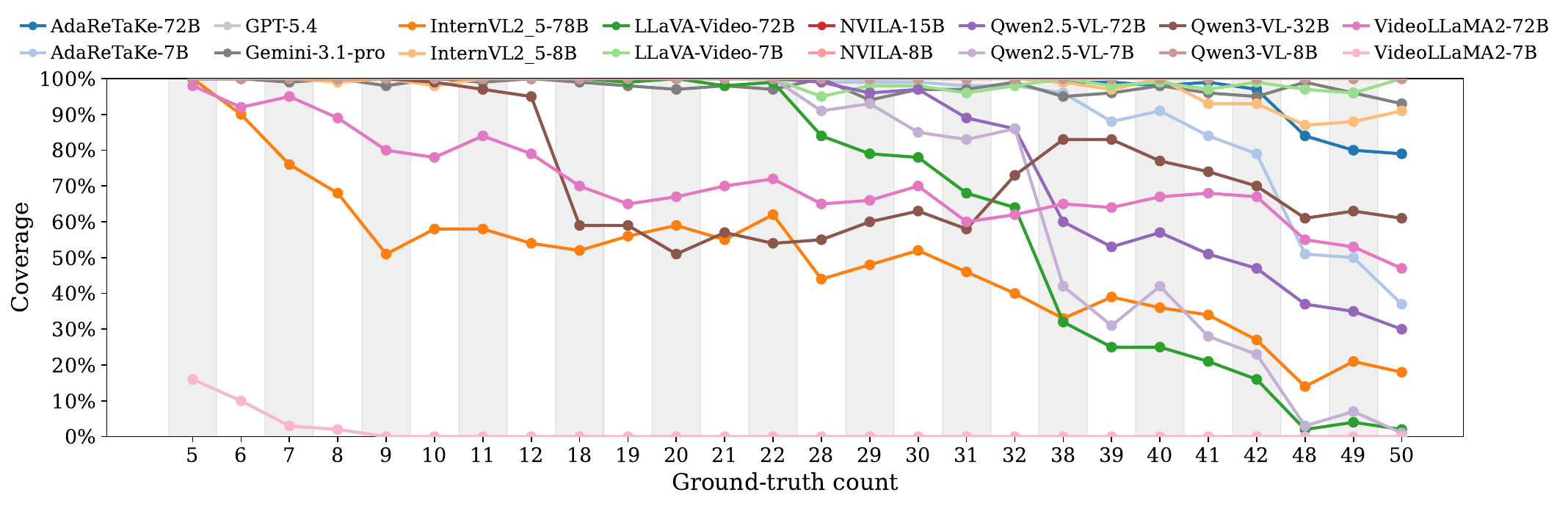}
    \caption{Model coverage by ground-truth count. Coverage is the proportion of videos for which the model produces a numerical answer.}
    \label{fig:coverage_curve}
\end{figure}

\begin{figure}[t]
    \centering
    \includegraphics[width=1.\textwidth]{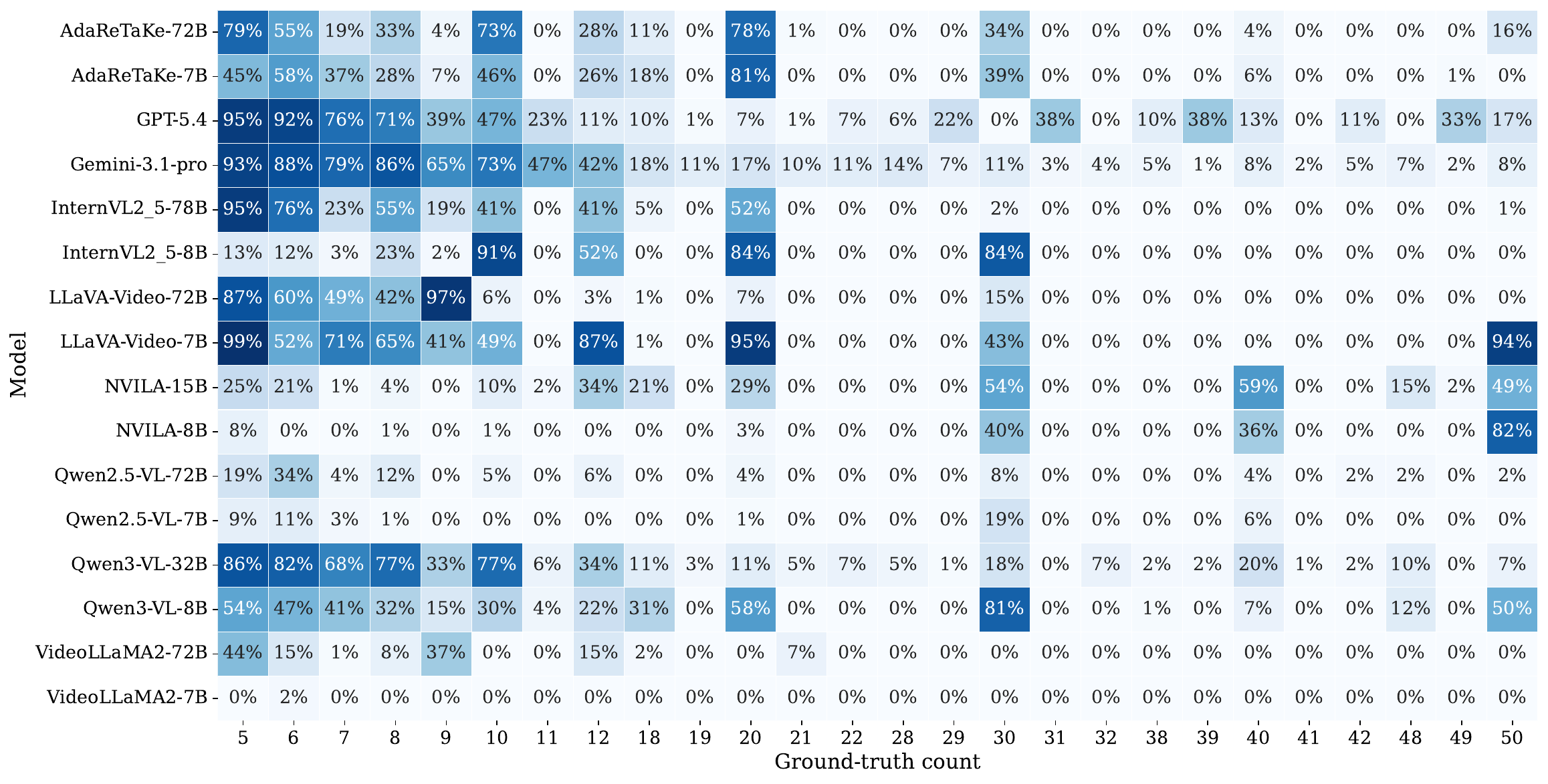}
    \caption{Exact Match accuracy by ground-truth count and model. 
    Each cell reports the percentage of videos for which the model's 
    predicted count equals the ground-truth exactly.}
    \label{fig:heatmap_em}
\end{figure}

The aggregate metrics of Section~\ref{sec:overall} hide a strong dependence of model behavior on the target cardinality. Figure~\ref{fig:coverage_curve} plots Coverage by ground-truth count, and Figure~\ref{fig:heatmap_em} reports the per-cardinality Exact Match.

\inlinehead{Coverage by cardinality.}
The compared models follow three behavior patterns in Figure~\ref{fig:coverage_curve}.
The first is full or near-full Coverage across the entire range.
Among the eight models with aggregate Coverage above $97\%$ in Table~\ref{tab:results}, five (NVILA-8B, NVILA-15B, Qwen3-VL-8B, GPT-5.4, and LLaVA-Video-7B) stay above $94\%$ on every cardinality, and two more (InternVL2.5-8B and Gemini-3.1-pro) drop slightly on the upper range but remain above $85\%$ throughout.
AdaReTaKe-72B is the borderline case: at near full Coverage up to $n=42$, it falls to $79\%$ at $n=50$.
The second pattern is a cardinality-dependent decline: seven models show this, six ending below $50\%$ Coverage at $n=50$ and Qwen3-VL-32B ending at $61\%$.
The onset varies widely from one model to the next: LLaVA-Video-72B remains near full Coverage until $n=22$ before falling to $2\%$ at $n=50$, while InternVL2.5-78B declines from $n=6$ onward, reaching $18\%$ at $n=50$.
The third pattern is a broad refusal to engage with the task: VideoLLaMA2-7B answers $16\%$ of videos at $n=5$ and essentially none above.

\inlinehead{Asymmetric dissociation of Coverage and Exact Match.}
Read together, Figures~\ref{fig:coverage_curve} and~\ref{fig:heatmap_em} show that the two metrics capture complementary failure modes (no answer versus wrong answers) and that this dissociation is asymmetric: EM can collapse while Coverage stays high, but no model preserves EM by selectively abstaining on hard cardinalities.
Qwen3-VL-8B answers every video at every cardinality while its EM drops to exactly zero at eleven of the twenty-six cardinalities: the model keeps answering, and answers wrong.
The opposite dissociation (Coverage dropping while EM is sustained by selective abstention on hard cases) does not occur in our evaluation.
LLaVA-Video-72B and InternVL2.5-78B both see Coverage and EM collapse jointly on the upper range: at $n=50$, LLaVA-Video-72B answers $2\%$ of the videos and none correctly, InternVL2.5-78B answers $18\%$ with EM $1\%$.
The aggregate Coverage reported in Table~\ref{tab:results} therefore captures a response policy rather than an epistemic calibration: high Coverage identifies models that always answer, but low Coverage does not identify models that abstain when they would have been wrong.

\inlinehead{Exact Match by cardinality.}
Counting larger sets is intuitively harder than counting small ones; one would therefore expect EM to decay with cardinality.
Figure~\ref{fig:heatmap_em} only partially supports this expectation.
A first group of models does decline as expected, with local irregularities but no large peaks: GPT-5.4 falls from $95\%$ at $n=5$ to $7\%$ at $n=20$ and stays below $20\%$ on most of the upper range, Gemini-3.1-pro falls from $93\%$ at $n=5$ to below $20\%$ for every $n \ge 18$, and similar drops are observed for InternVL2.5-78B, LLaVA-Video-72B, and Qwen3-VL-32B.
The remaining models are non-monotonic, with pronounced peaks at specific cardinalities and near-zero accuracy at adjacent counts: LLaVA-Video-7B reaches EM $95\%$ at $n=20$, $43\%$ at $n=30$, and $94\%$ at $n=50$; AdaReTaKe-72B peaks at $73\%$ at $n=10$ and $78\%$ at $n=20$; InternVL2.5-8B reaches EM between $84\%$ and $91\%$ at $n \in \{10, 20, 30\}$ and collapses to zero at every intermediate cardinality except $n=12$ ($52\%$); Qwen3-VL-8B peaks at $81\%$ at $n=30$ with markedly lower values on either side.
The two NVILA variants share this peak-and-collapse structure, with peaks on the upper anchors $\{30, 40, 50\}$, producing an apparent upward trend when only endpoints are compared.

EM therefore does not follow one difficulty-driven trajectory: models with comparable aggregate EM either decline monotonically or oscillate along the cardinality axis. Across the non-monotonic group, the peak-and-collapse structure dominates: isolated peaks at specific cardinalities, with near-zero accuracy at adjacent counts.
AdaReTaKe-7B illustrates this: $81\%$ at $n=20$, $39\%$ at $n=30$, near zero on adjacent cardinalities. Peak locations vary across models, but the structure itself is shared.
It cannot be explained by cardinality-as-difficulty alone, which Section~\ref{sec:bias} addresses by analysing prediction distributions.

\subsection{Distributional Biases in Model Predictions}
\label{sec:bias}

\begin{figure}[!t]
    \centering

    \begin{tabular}{c}
        \includegraphics[width=\textwidth]{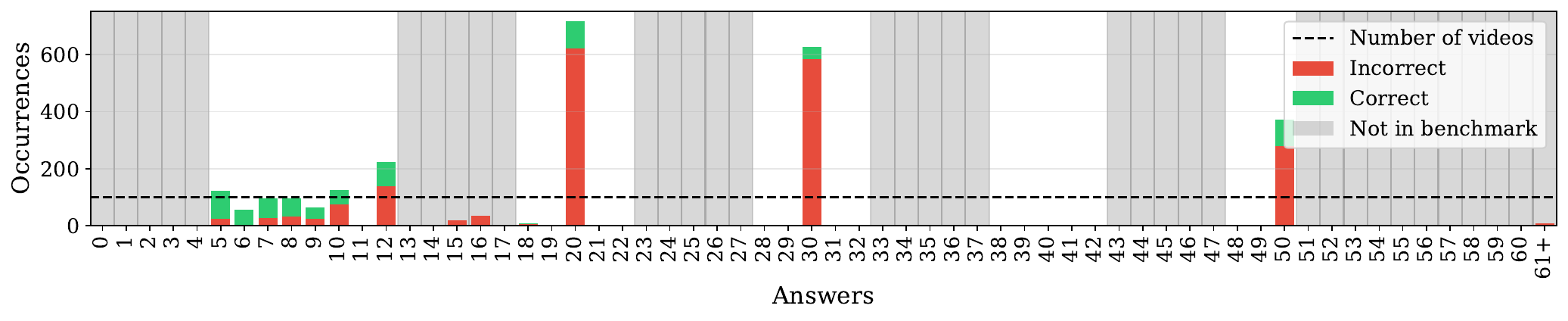} \\
        (a) LLaVA-Video-7B \\[0.15em]

        \includegraphics[width=\textwidth]{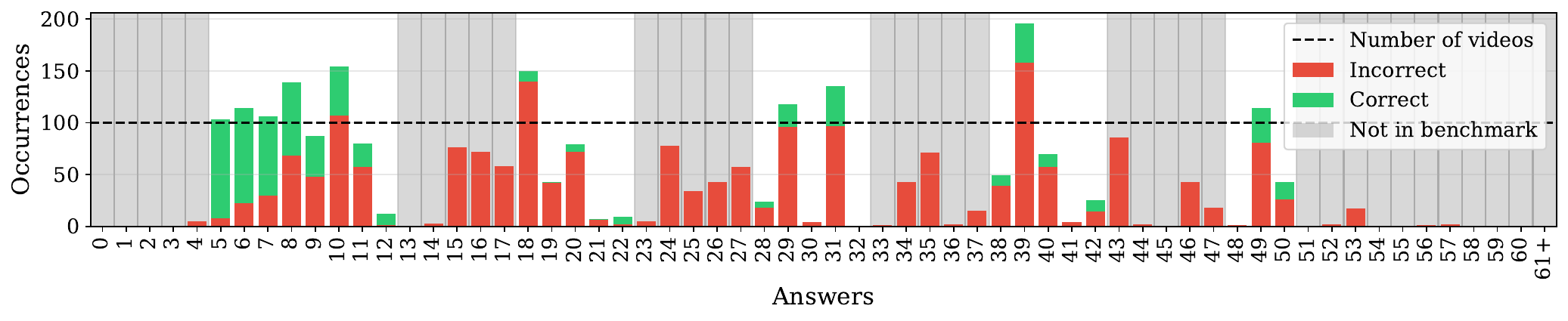} \\
        (b) GPT-5.4 \\[0.15em]%

        \includegraphics[width=\textwidth]{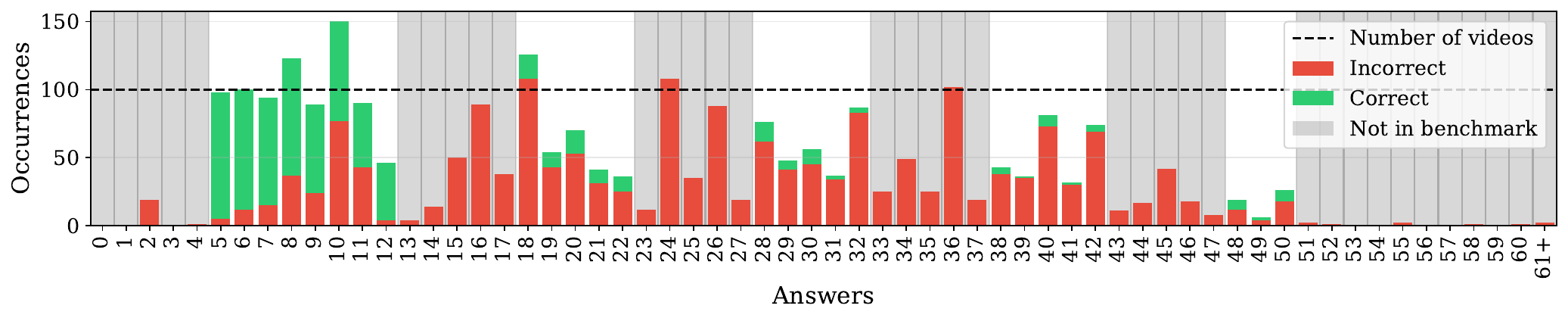} \\
        (c) Gemini-3.1-pro
    \end{tabular}

    \caption{Distributions of predicted counts on CountVideoBench. Each bar reports the number of times a given integer value is predicted by the model across the $2{,}600$ videos. (a) LLaVA-Video-7B. (b) GPT-5.4. (c) Gemini-3.1-pro.}
    \label{fig:biais_barplots}
\end{figure}

\inlinehead{EM peaks coincide with preferred values.} The peak-and-collapse structure of Exact Match noted at the end of Section~\ref{sec:cardinality} admits a natural reading: the model would be locally competent, accurate at some cardinalities and weak at adjacent ones.
The distribution of predictions rules out this reading. LLaVA-Video-7B provides the clearest case.
Figure~\ref{fig:biais_barplots}(a) shows the histogram of its predicted counts across the $2{,}600$ videos: sixteen integer values appear at all, and five of them concentrate $80\%$ of the mass: $20$ (predicted $716$ times), $30$ ($628$), $50$ ($373$), $12$ ($224$), and $10$ ($124$).
The Exact Match peaks of Figure~\ref{fig:heatmap_em} line up with four of these five preferred values: EM is $95\%$ at $n=20$, $43\%$ at $n=30$, $94\%$ at $n=50$, and $87\%$ at $n=12$.
The fifth, $10$, behaves differently: the value $10$ is predicted $124$ times across the benchmark, but only $49$ of those predictions occur at $n=10$, so EM at $n=10$ stays at $49\%$.
At every cardinality outside the preferred set the model falls back on the closest preferred value, which is incorrect by construction; the value $40$ is never predicted across the entire benchmark, and EM at $n=40$ is consequently $0\%$.
The Exact Match peaks therefore mark coincidences between a small, model-specific set of preferred answers and the ground-truth value, rather than cardinalities at which the model counts well.
The same peak-and-collapse structure appears in Figure~\ref{fig:heatmap_em} for AdaReTaKe-7B, InternVL2.5-8B, and NVILA-8B, each with its own preferred set.

\inlinehead{Perplexity reveals a narrow effective vocabulary.}
Table~\ref{tab:results} quantifies this concentration of mass across all evaluated models.
The Oracle row gives the reference values of a model that would predict the ground truth on every video: support equal to the twenty-six benchmark cardinalities, perplexity equal to twenty-six as well.
VideoLLaMA2-7B sits at the opposite end of the table ($|\mathcal{S}|=5$, $\mathrm{PP}=1.1$); these values reflect its near-zero coverage rather than a value-dependent bias.
Twelve of the fifteen non-degenerate models stay well below the Oracle reference, with support between sixteen and thirty-two distinct integer answers and perplexity below $18$.
Four have perplexity below $10$: InternVL2.5-78B ($5.9$), LLaVA-Video-7B ($8.0$), VideoLLaMA2-72B ($8.9$), and InternVL2.5-8B ($9.4$).
These models effectively answer the benchmark with the equivalent of fewer than ten equally-frequent values, against twenty-six target cardinalities. The gap between $|\mathcal{S}|$ and $\mathrm{PP}$ is also informative: VideoLLaMA2-72B uses twenty-five distinct values but has perplexity $8.9$, indicating that the bulk of its mass concentrates on a small subset of its support.
Three models stand apart with substantially larger diversity: Gemini-3.1-pro ($|\mathcal{S}|=55$, $\mathrm{PP}=38.9$), GPT-5.4 ($48$, $30.6$), and Qwen3-VL-32B ($83$, $27.8$).
All three exceed the Oracle perplexity: their predictions span integer values outside the twenty-six benchmark cardinalities.

\inlinehead{High perplexity does not imply uniformity.}
The diversity reported in Table~\ref{tab:results} does not imply uniformity.
Figure~\ref{fig:biais_barplots}(b)--(c) plots the histograms of predicted counts for the two proprietary models, which sit at the top of the perplexity column.
Both distributions exhibit pronounced local irregularities.
For GPT-5.4, the largest local excess across its support is observed at value $39$, with $+126$ predictions above its neighbours, and the largest local deficit at value $30$, a benchmark anchor cardinality, with $-114$ predictions below its flanking peaks. Gemini-3.1-pro shows the same qualitative pattern with smaller amplitudes: maximum local excess of $+53$ at value $36$, maximum local deficit of $-57$ at value $27$.
The aggregate perplexity therefore captures only the global spread of the prediction distribution, not its local structure: even a model with $\mathrm{PP} > 30$ can concentrate its predictions on isolated values and avoid others, including benchmark anchor cardinalities.
The same kind of local structure is visible in the histograms of the remaining models, shown in Appendix Figs.~\ref{fig:biais_barplots_annexe_1}--\ref{fig:biais_barplots_annexe_8}.

The biases documented in this section are visible only because every cardinality in $[5, 50]$ contributes the same number of videos. On an imbalanced benchmark, the per-cardinality peaks, the Coverage--EM dissociation, and the local excesses and deficits of the prediction histograms would all be absorbed into the dominant counts, conflating response policy, output diversity, and value-dependent biases into a single aggregate score. The structure persists across three prompt formulations on three models and is not explained by object class, though absolute accuracy varies with both (Appendices~\ref{app:prompts} and~\ref{app:classes}).

\section{Conclusion}
\label{sec:conclusion}

This paper introduced CountVideoBench, a balanced benchmark of $2{,}600$ synthetic videos covering $26$ cardinalities in the range $[5, 50]$ with $100$ videos per cardinality, organized in an anchor-neighborhood structure that combines a broad cardinality range with the local resolution needed to compare neighboring counts.
Sixteen video-language models from nine architecture families were considered.
A multi-criteria evaluation is proposed using a single zero-shot protocol.
Three findings emerged from the analysis.
No model exceeds $27.6\%$ Exact Match on a task involving a single object class, static instances, and a controlled camera motion: counting remains unresolved even after recognition and object motion have been removed as confounds.
The dependence of model behavior on cardinality is not a simple difficulty-driven decay (trajectories along the cardinality axis vary across models, either declining monotonically or following a non-monotonic peak-and-collapse structure). Coverage degradation, when it occurs, joins this decay rather than offsetting it through selective abstention.
Every model concentrates its predictions on its own set of preferred integer values, with accuracy peaks coinciding with these values and collapsing at adjacent counts.
In most models, predictions are confined to a small set of values; in those with the broadest output distributions, the concentration appears instead as pronounced local irregularities.

These observations could not be characterized on the cardinality distributions of existing Video-QA benchmarks, where aggregate scores conflate response policy, output diversity, and value-dependent biases into a single number.
The balanced design of CountVideoBench makes the underlying structure visible without resolving its origin.
The design suggests several extensions.
Adding distractor object classes would measure the cost of reintroducing recognition under the same counting protocol; introducing object motion would extend the analysis to temporal tracking and accumulation.
The balanced-distribution principle could also be applied beyond the synthetic regime, to natural video, and transposed to other VideoQA dimensions where similar value-dependent biases may remain hidden behind imbalanced distributions: attribute recognition, action enumeration, and temporal estimation.

\section*{Acknowledgement}

We are grateful to Louis Airale for his sustained involvement in this work, through many discussions and detailed comments on earlier drafts.
We also thank the PAGODA team, whose computational resources and constant availability made this work possible.

\bibliography{egbib}

\appendix

\newpage

\section{Count Distributions in Existing Benchmarks}
\label{app:dist}

\begin{figure}[H]
    \centering
    \includegraphics[width=0.65\linewidth]{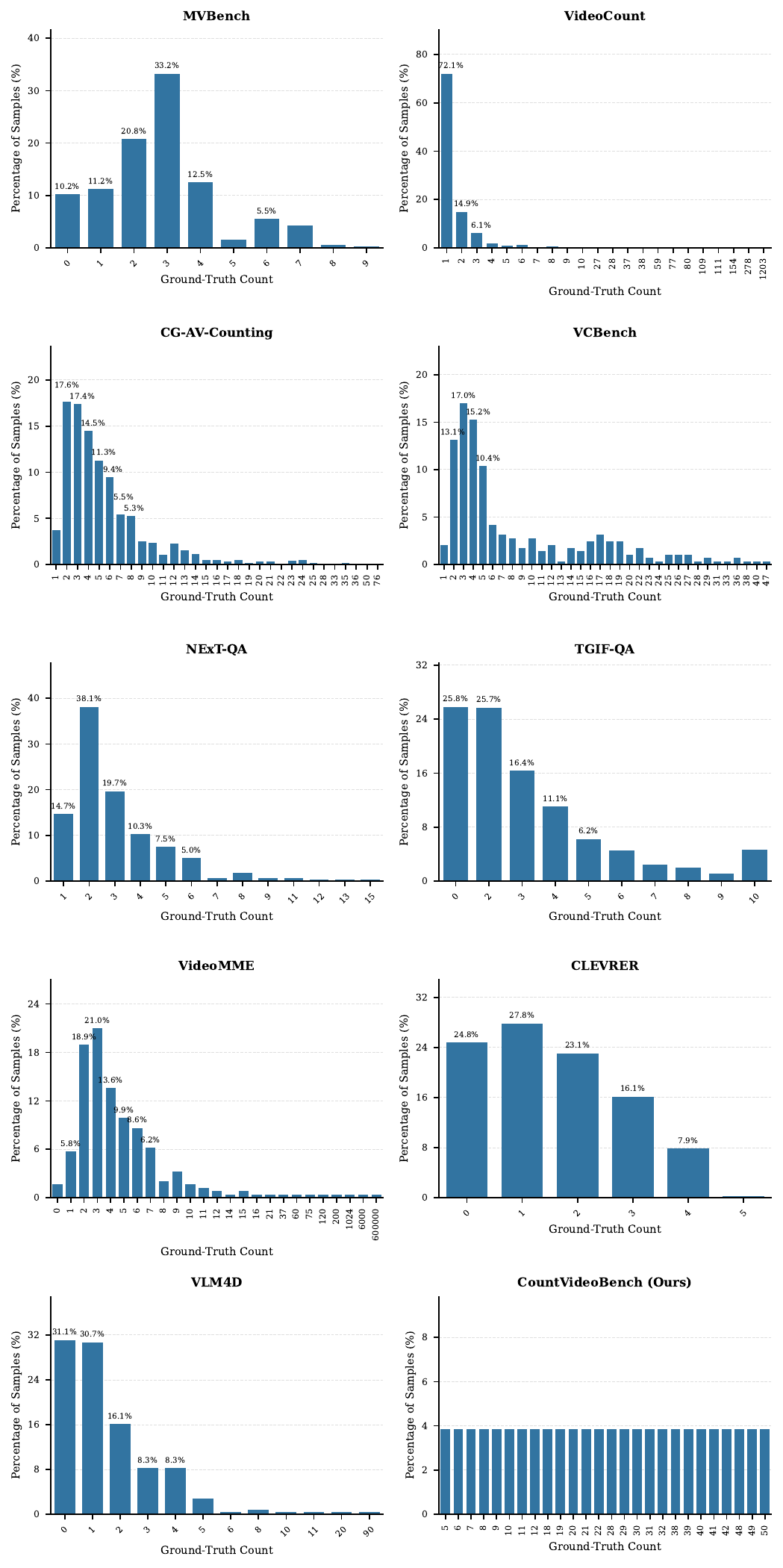}
    \caption{Ground-truth count distributions in the representative video
    benchmarks with counting questions. Each panel shows the number of videos at each ground-truth
    count.}
    \label{fig:dist_existing_annexe}
\end{figure}

\section{Generation Pipeline Parameters}
\label{app:pipeline}

Table~\ref{tab:pipeline_params} reports the parameters that fully determine the generated scenes.
All lengths are expressed in Blender units; the reference scale is the cube edge, which measures $2$ units.
Every video is determined by a single random seed, which is used to seed both the Python and NumPy generators.

\begin{table}[!t]
\centering
\scriptsize
\caption{Generation parameters. $\mathcal{U}(a,b)$ denotes a uniform draw on $[a,b]$. The polar angle is drawn subject to the placement constraint described below, which induces the effective distribution reported in the last row of the camera block.}
\label{tab:pipeline_params}
\begin{tabular}{ll}
\toprule
Parameter & Value \\
\midrule
\multicolumn{2}{l}{\emph{Rendering}} \\
Engine & Cycles, GPU, light tree enabled \\
Samples & $32$ max, adaptive sampling, noise threshold $0.2$ \\
Light paths & $4$ total (diffuse $2$, glossy $2$, transmission $4$, volume $2$) \\
Clamping & direct and indirect $5.0$ \\
Denoising & disabled for final renders \\
Colour management & display sRGB, view transform Filmic, look \textsc{high contrast} \\
 & exposure $0$, gamma $1$, white balance off \\
Resolution & $1920 \times 1080$ at $100\%$ scale \\
Frame rate & $30$ fps \\
Settling window & frames $1$--$20$ (not rendered) \\
Rendered window & frames $20$--$170$ ($150$ frames, $5$ s) \\
Renderer setup & Poly Render Pro v$3.2$ (GPL), \textsc{fast} Cycles preset \\
\midrule
\multicolumn{2}{l}{\emph{Camera}} \\
Focal length & $50$ mm \\
Sensor width & $36$ mm, sensor fit \textsc{auto} \\
Field of view & $39.60^{\circ}$ horizontal, $22.90^{\circ}$ vertical \\
Position & fixed for the whole clip \\
Distance to origin & $\mathcal{U}(32, 37)$ \\
Azimuth & $\mathcal{U}(0^{\circ}, 360^{\circ})$ \\
Polar angle (drawn) & $\mathcal{U}(0^{\circ}, 65^{\circ})$ from the vertical axis \\
Polar angle (effective) & $[58.2^{\circ}, 65.0^{\circ}]$, mean $63.1^{\circ}$, std $1.4^{\circ}$ \\
\midrule
\multicolumn{2}{l}{\emph{Camera orientation}} \\
Keyframes & frames $1, 50, 100, 150, 200$ \\
Interpolation & B\'ezier, auto-clamped handles \\
Elevation offset & $\mathcal{U}(-1^{\circ}, +5^{\circ})$ per keyframe \\
Azimuth offset & $\mathcal{U}(-5^{\circ}, +5^{\circ})$ per keyframe \\
\midrule
\multicolumn{2}{l}{\emph{Objects}} \\
Cube & edge $2$, bevel width $0.1$, $100$ segments \\
Sphere & radius $1$, UV sphere $64 \times 32$, smooth shading \\
Vertical placement & bounding-box base resting on the plane $z = 0$ \\
Planar jitter & $\pm 0.30$ (cubes), $\pm 0.25$ (spheres) \\
Yaw & $\mathcal{U}(0, 2\pi)$ \\
Material & procedural, instanced per object \\
\quad noise realisation & $\mathcal{U}(-20, 20)$ \\
\quad mapping orientation & $\mathcal{U}(-\pi, \pi)^{3}$ \\
\midrule
\multicolumn{2}{l}{\emph{Placement}} \\
Admissible region & camera frustum $\cap$ plane $z = 0$, scaled by $0.9$ \\
Minimum separation & $3$ between object centres, in the plane \\
Sampling & uniform rejection sampling \\
Positions generated & $60$ per scene, first $N$ retained \\
\midrule
\multicolumn{2}{l}{\emph{Physics}} \\
Rigid body & active, box collision shape \\
Mass & $1.0 \times \text{object size}$ \\
Friction / restitution & $1.0$ / $0.3$ \\
Linear / angular damping & $0.6$ / $0.6$ \\
\midrule
\multicolumn{2}{l}{\emph{Lighting}} \\
Sources & five area lights, plus an HDRI environment map for indirect light \\
Arrangement & one overhead, four at the scene border angled inward \\
Environment map & \emph{Limpopo Golf Course} (Poly Haven, CC0), strength $0.4$ \\
Orientation & fixed; not visible in render (occluded by the backdrop) \\
Variation & none; the lighting setup is identical across all videos \\
\bottomrule
\end{tabular}
\vspace{0.5em}
\begin{minipage}{\linewidth}
{\footnotesize\emph{Assets.} The environment map \emph{Limpopo Golf Course} is from
Poly Haven (CC0). Object and background materials are procedural BlenderKit assets
used under their Royalty Free licence, which permits their use in rendered output
but not their redistribution; they are therefore not included in the released code,
which lists them as external dependencies.\par}
\end{minipage}
\end{table}

\inlinehead{Object placement.}
Object positions are sampled in the plane $z = 0$ rather than in a volume.
The admissible region is the intersection of the camera frustum with that plane, scaled by a factor $0.9$ about its centroid so that no object is placed exactly at the edge of the initial field of view.
Positions are drawn by uniform rejection sampling within the bounding box of this region, retaining a candidate only if it lies inside the region and is at least $3$ units away from every previously accepted position.
The sampler always produces $60$ valid positions and retains the first $N$, so the packing problem is identical at every cardinality: the objects of a video with $N = 5$ occupy the same region, under the same separation constraint, as those of a video with $N = 50$.
Increasing difficulty with cardinality therefore does not arise from a tightening of the scene geometry.

\inlinehead{Camera pose.}
The camera is placed on a sphere centred on the origin and remains at that position for the whole clip; only its orientation varies.
The polar angle is drawn uniformly on $[0^{\circ}, 65^{\circ}]$ subject to the constraint that the resulting admissible region accommodate the $60$ required positions; poses that fail this constraint are redrawn.
The constraint is binding, since the ground area visible to the camera grows with obliquity: at the mean camera distance, the admissible area is close to $900$ units\textsuperscript{2} at $63^{\circ}$, $680$ at $58^{\circ}$ and $440$ at $45^{\circ}$, whereas random sequential adsorption of $60$ discs of radius $1.5$ saturates at an area on the order of $750$--$800$ units\textsuperscript{2}.
The admissible area drops below that range near $58^{\circ}$--$60^{\circ}$, which is consistent with the observed minimum of $58.2^{\circ}$.
The effective distribution of the polar angle is therefore concentrated on $[58.2^{\circ}, 65.0^{\circ}]$, with mean $63.1^{\circ}$ and standard deviation $1.4^{\circ}$, placing the camera at approximately $27^{\circ}$ above the ground plane.
Viewpoint variation across the benchmark is thus essentially azimuthal, the azimuth being unconstrained by rotational symmetry; camera elevation is close to constant and is not a source of variation between videos.

\inlinehead{Separation and non-overlap.}
The minimum separation of $3$ units is enforced on the sampled positions, before the planar jitter of $\pm 0.30$ (cubes) or $\pm 0.25$ (spheres) is applied.
For spheres of radius $1$, the worst-case centre distance after jitter remains above $2.29$, so contact is impossible.
For cubes of edge $2$ with a uniformly drawn yaw, the worst case brings the centre distance down to $2.15$ while corner-to-corner contact requires $2.83$, so a residual interpenetration is possible in principle.
These cases are resolved by the rigid-body solver: the physics cache is baked before rendering, and objects come to rest during the settling window that precedes the recorded frames.
The rendered clips therefore contain no object motion.

\inlinehead{Visibility measurement.}
An object counts as inside the frame at a given frame if the projection of its origin falls within $[0,1]^{2}$ in normalised camera coordinates with positive depth, which accounts for focal length, sensor size and render aspect ratio.
Applying this criterion to every object at every rendered frame yields the statistics reported in Section~\ref{sec:pipeline}: in $83.0\%$ of the $2{,}600$ videos, at least one object falls outside the frame at some point, and averaged over all videos, $3.29$ objects per video do so.
Relative to the mean cardinality of the benchmark ($25.6$), this amounts to roughly $13\%$ of the objects of a typical video leaving the frame at some point during the clip.

\section{Human Baseline}
\label{app:human}

Section~\ref{sec:overall} reports that no model exceeds $27.6\%$ Exact Match on CountVideoBench. Two readings of this number are possible: either the models fail at a task that is solvable, or the scenes themselves are too dense or too briefly observed to be counted exactly. We ran a human study to separate them.

\inlinehead{Participants.} Twenty volunteers took part, without compensation.
They were told only to count the objects of the specified class as accurately as they could; the purpose of the study, the cardinality range, and the existence of a model comparison were not disclosed. Participation could be interrupted at any point.
All twenty completed the full protocol, giving a fully crossed design in which every participant answers every video, so that participant ability and video difficulty are not confounded.

\inlinehead{Stimuli.} We sampled $104$ videos, four per cardinality across the $26$ values of the benchmark, drawn at random within each cardinality under a constraint keeping the ten object classes close to balanced overall.
Participants viewed exactly the ten uniformly sampled frames given to the models throughout the paper, presented as a playable sequence; the full $150$-frame clip was not available to them.
Playback was unconstrained: participants could pause, step back, and take as long as they wished. Each video was answered with a single integer, and no answer could be skipped.

\inlinehead{Aggregation.} Exact Match is computed over the $2{,}080$ responses ($20$ participants $\times$ $104$ videos). Support size and perplexity are computed per participant over their own responses and then averaged.
All rows of Table~\ref{tab:human}, human and model alike, are computed over the same $104$ videos.
We additionally report a \emph{majority-vote} aggregate, in which each video receives the modal answer of the twenty participants.

\inlinehead{Accuracy.} Mean human Exact Match reaches $90.8\%$, against $26.9\%$ for the best models on this subset (Gemini-3.1-pro and LLaVA-Video-7B).
Per-participant accuracy ranges from $71.2\%$ to $100.0\%$ (median $91.8$, standard deviation $8.1$), so the gap is not carried by a subset of strong counters: the weakest participant remains far above the best model. Two participants answered all $104$ videos correctly, and the majority-vote aggregate is exact on every video. No scene in the subset is therefore intrinsically ambiguous, and the low model scores reflect limitations of current approaches rather than uncountable inputs.

\begin{figure}[H]
    \centering
    \includegraphics[width=1.\textwidth]{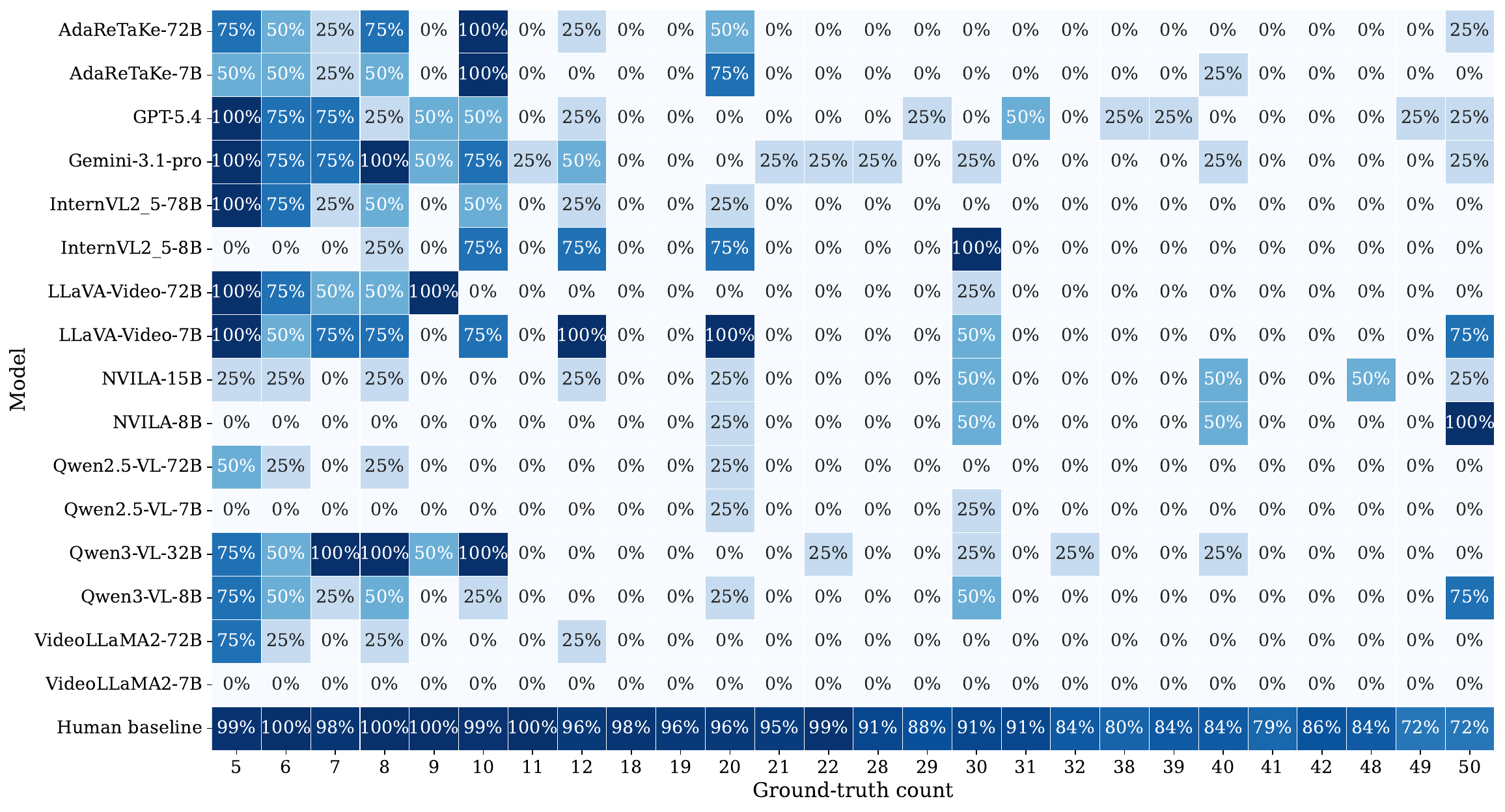}
    \caption{Exact Match by ground-truth cardinality on the $104$-video human-study subset, for all evaluated models and for the twenty human participants (bottom row). Each model cell aggregates four videos and therefore takes values in $\{0, 25, 50, 75, 100\}\%$; each human cell aggregates $80$ responses. The human row declines steadily from $99\%$ at the smallest counts to $74\%$ at $n=50$, without the isolated peaks and adjacent collapses visible across the model rows.
}
    \label{fig:human_cardinality}
\end{figure}

\inlinehead{Behavior across cardinality.} Figure~\ref{fig:human_cardinality} shows human accuracy against ground-truth count.
It declines gradually with cardinality, from around $99\%$ at the smallest counts to $92\%$ at $n=20$, $85\%$ at $n=40$ and $74\%$ at $n=50$.
High cardinalities are therefore harder for humans as well, but no value falls below $74\%$: the isolated peaks and adjacent collapses
characterized in Section~\ref{sec:cardinality} have no counterpart in the human profile.

\inlinehead{Prediction distribution.} Humans use $28.7$ distinct values on average with a perplexity of $27.3$, against $26.0$ for the Oracle: their errors spread almost uniformly around the true value. Models at a comparable support behave differently. Gemini-3.1-pro reaches $|\mathcal{S}| = 39$ for $\mathrm{PP} = 33.1$, and LLaVA-Video-7B concentrates $12$ values into $\mathrm{PP} = 7.6$. The preferred-value concentration reported in Section~\ref{sec:bias} is thus not a property of the task: respondents that also make errors on these videos do not exhibit it.

\inlinehead{Limitations.} The study covers $104$ of the $2{,}600$ videos, and human viewing time was unconstrained whereas models process the ten frames in a single pass.
The comparison therefore establishes that the scenes are countable under unrestricted inspection; it is not a matched-effort comparison.

\begin{table}[H]
\centering
\caption{Human baseline and model performance on the $104$-video subset. Human
rows report the mean over $20$ participants; the majority-vote row aggregates the
modal answer per video. Support size and perplexity are per-respondent averages.}
\label{tab:human}
\begin{tabular}{lccccc}
\toprule
Respondent & EM & Acc@10\% & Coverage & $|\mathcal{S}|$ & $\mathrm{PP}$ \\
\midrule
AdaReTaKe-7B                   &  14.4 &  30.8 &  90.4 & 14.0 &  9.7 \\
AdaReTaKe-72B                  &  16.3 &  40.4 &  96.2 & 18.0 & 13.3 \\
InternVL2.5-8B                 &  13.5 &  36.5 &  97.1 & 12.0 &  9.6 \\
InternVL2.5-78B                &  13.5 &  19.2 &  44.2 & 10.0 &  5.0 \\
LLaVA-Video-7B                 &  26.9 &  56.7 &  99.0 & 12.0 &  7.6 \\
LLaVA-Video-72B                &  15.4 &  31.7 &  67.3 & 17.0 &  9.8 \\
NVILA-8B                       &   8.7 &  26.9 & 100.0 & 17.0 & 10.5 \\
NVILA-15B                      &  11.5 &  42.3 & 100.0 & 19.0 & 13.3 \\
Qwen2.5-VL-7B                  &   1.9 &   4.8 &  72.1 & 17.0 & 11.0 \\
Qwen2.5-VL-72B                 &   4.8 &  11.5 &  80.8 & 19.0 & 14.4 \\
Qwen3-VL-8B                    &  14.4 &  43.3 & 100.0 & 23.0 & 17.4 \\
Qwen3-VL-32B                   &  22.1 &  34.6 &  69.2 & 35.0 & 18.1 \\
VideoLLaMA2-7B                 &   0.0 &   0.0 &   1.0 &  1.0 &  1.1 \\
VideoLLaMA2-72B                &   5.8 &   6.7 &  74.0 & 16.0 &  9.4 \\
GPT-5.4                        &  22.1 &  66.3 & 100.0 & 28.0 & 23.6 \\
Gemini-3.1-pro                 &  26.9 &  57.7 &  98.1 & 39.0 & 33.1 \\
\midrule
Human baseline                 &  \textbf{90.8} & \textbf{98.8} & 100.0 & 28.7 & 27.3 \\
Human (majority vote)          & 100.0 & 100.0 & 100.0 & 26.0 & 26.0 \\
Oracle                         & 100.0 & 100.0 & 100.0 & 26.0 & 26.0 \\
\bottomrule
\end{tabular}
\end{table}

\section{Prediction Distributions by Model}
\label{app:barplots}

\begin{figure}[H]
    \centering

    \begin{tabular}{c}
        \includegraphics[width=\textwidth]{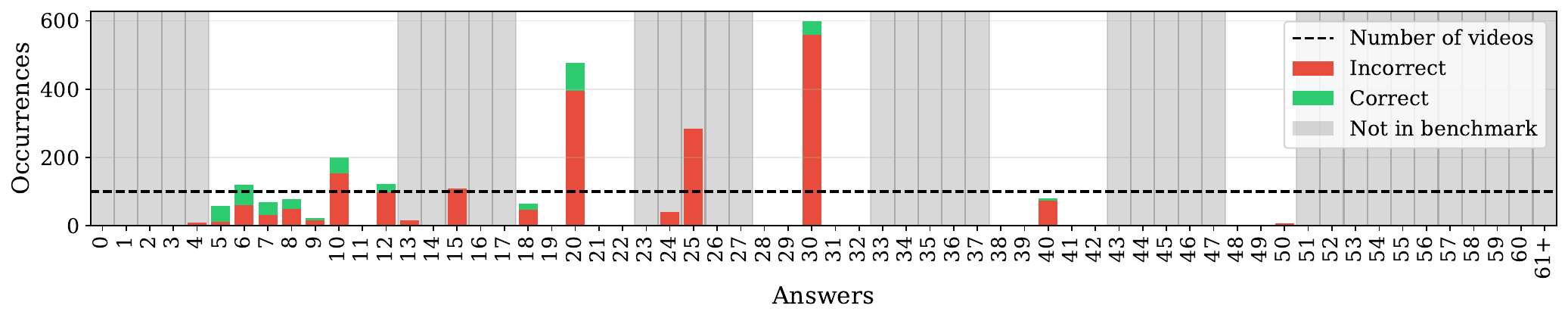} \\
        (a) AdaReTaKe-7B \\[0.15em]

        \includegraphics[width=\textwidth]{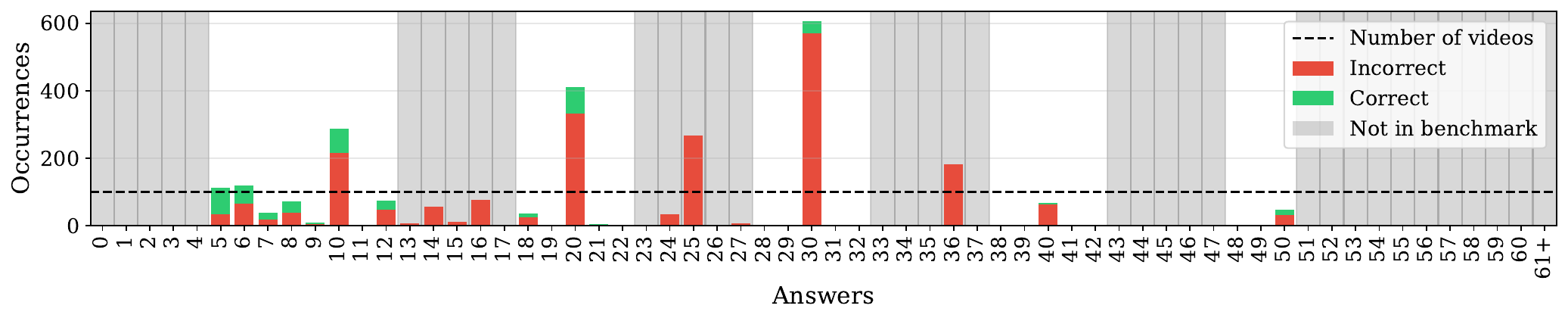} \\
        (b) AdaReTaKe-72B \\[0.15em]
    \end{tabular}

    \caption{Distributions of predicted counts on CountVideoBench. Each bar reports the number of times a given integer value is  predicted by the model across the $2{,}600$ videos.}
    \label{fig:biais_barplots_annexe_1}
\end{figure}

\begin{figure}[H]
    \centering

    \begin{tabular}{c}
        \includegraphics[width=\textwidth]{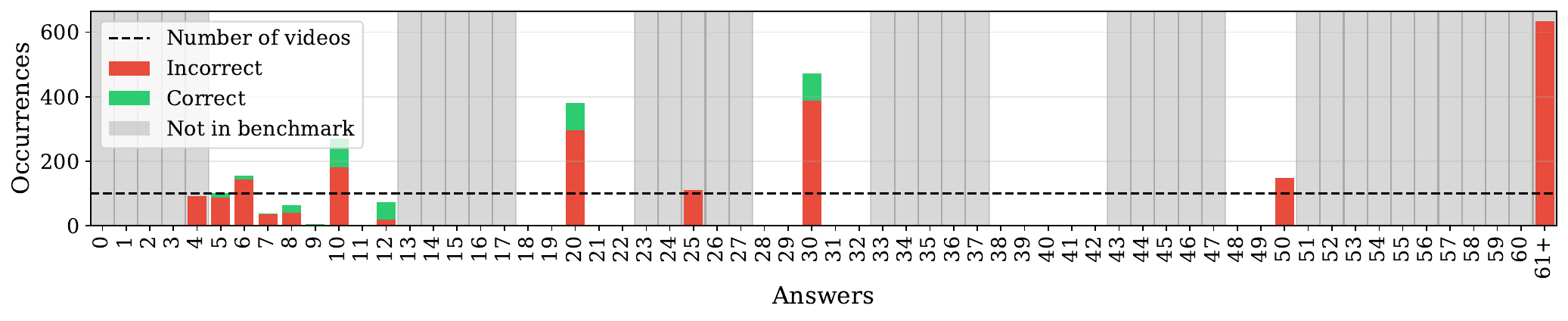} \\
        (c) InternVL2.5-8B \\[0.15em]

        \includegraphics[width=\textwidth]{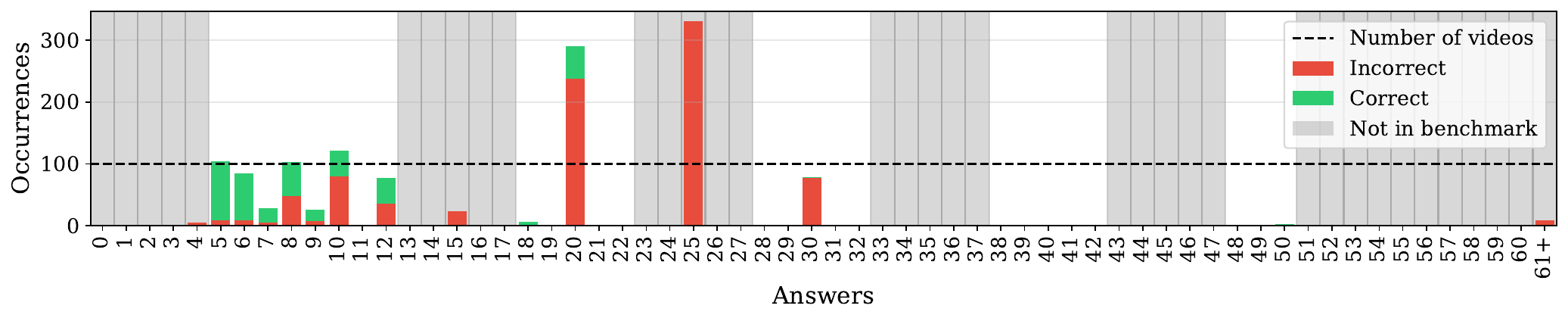} \\
        (d) InternVL2.5-78B \\[0.15em]
    \end{tabular}

    \caption{Distributions of predicted counts on CountVideoBench. Each bar reports the number of times a given integer value is  predicted by the model across the $2{,}600$ videos.}
    \label{fig:biais_barplots_annexe_2}
\end{figure}

\begin{figure}[H]
    \centering

    \begin{tabular}{c}
        \includegraphics[width=\textwidth]{captions/biais/LLaVA-Video-7B_bias_barplot_shorter.pdf} \\
        (e) LLaVA-Video-7B \\[0.15em]

        \includegraphics[width=\textwidth]{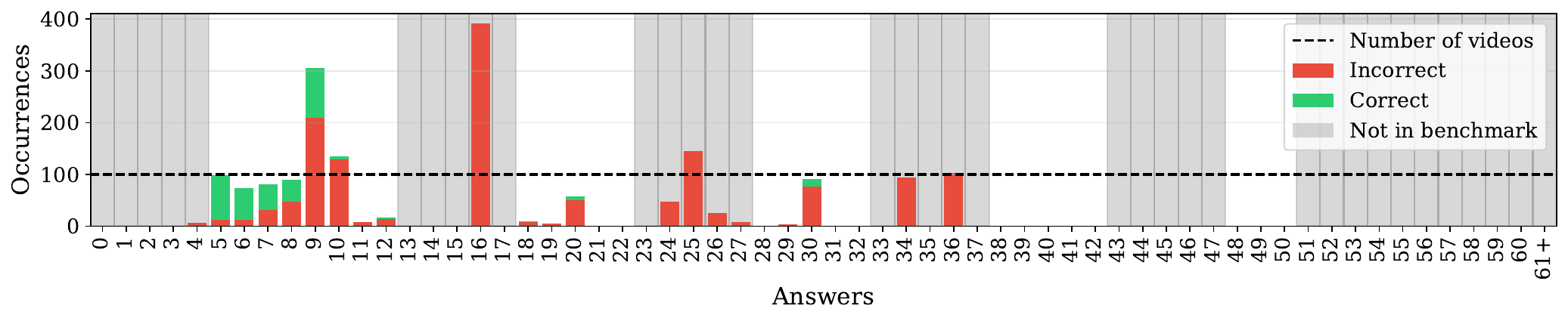} \\
        (f) LLaVA-Video-72B \\[0.15em]
    \end{tabular}

    \caption{Distributions of predicted counts on CountVideoBench. Each bar reports the number of times a given integer value is  predicted by the model across the $2{,}600$ videos.}
    \label{fig:biais_barplots_annexe_3}
\end{figure}

\begin{figure}[H]
    \centering

    \begin{tabular}{c}
        \includegraphics[width=\textwidth]{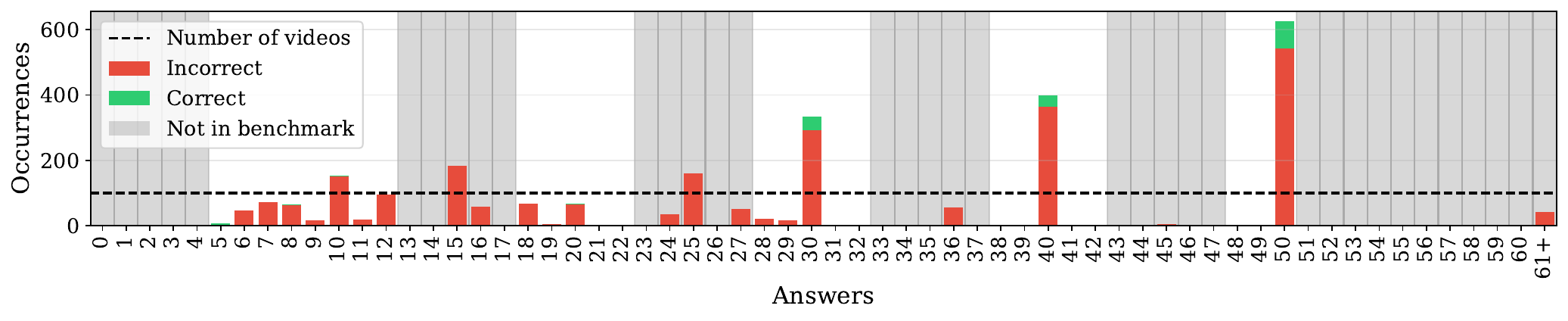} \\
        (g) NVILA-8B \\[0.15em]

        \includegraphics[width=\textwidth]{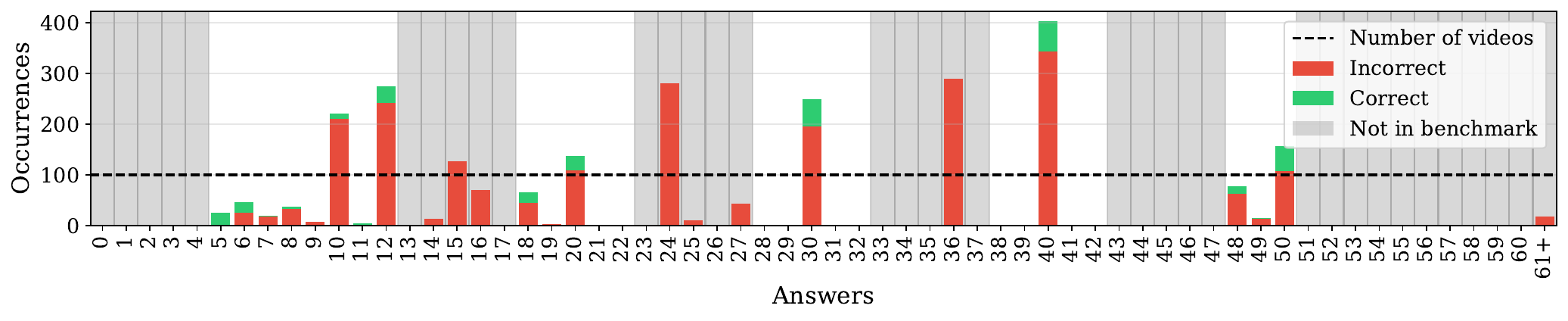} \\
        (h) NVILA-15B \\[0.15em]
    \end{tabular}

    \caption{Distributions of predicted counts on CountVideoBench. Each bar reports the number of times a given integer value is  predicted by the model across the $2{,}600$ videos.}
    \label{fig:biais_barplots_annexe_4}
\end{figure}

\begin{figure}[H]
    \centering

    \begin{tabular}{c}
        \includegraphics[width=\textwidth]{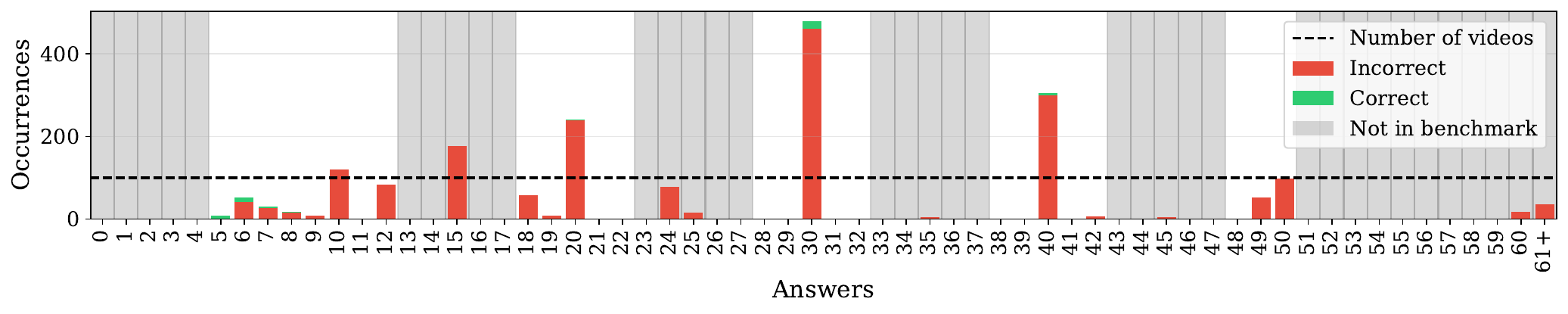} \\
        (i) Qwen2.5-VL-7B \\[0.15em]

        \includegraphics[width=\textwidth]{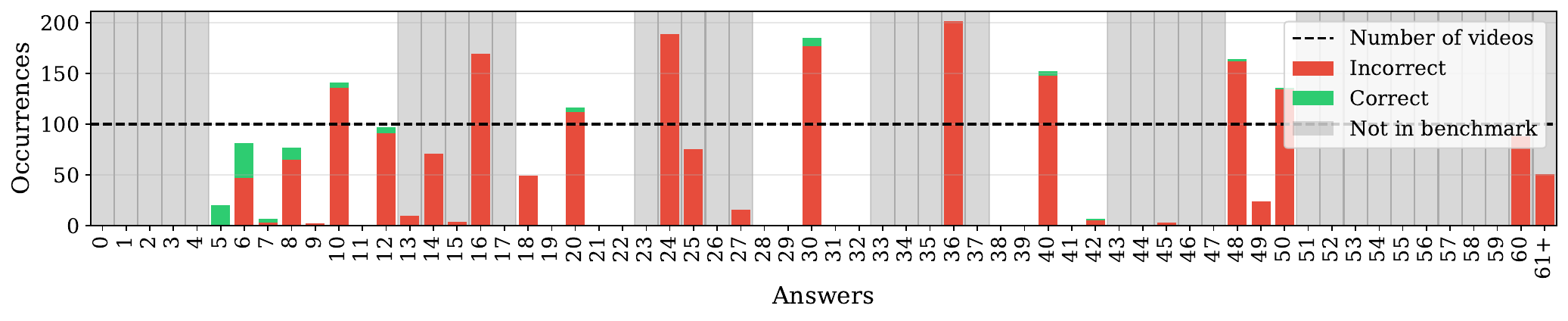} \\
        (j) Qwen2.5-VL-72B \\[0.15em]
    \end{tabular}

    \caption{Distributions of predicted counts on CountVideoBench. Each bar reports the number of times a given integer value is  predicted by the model across the $2{,}600$ videos.}
    \label{fig:biais_barplots_annexe_5}
\end{figure}

\begin{figure}[H]
    \centering

    \begin{tabular}{c}
        \includegraphics[width=\textwidth]{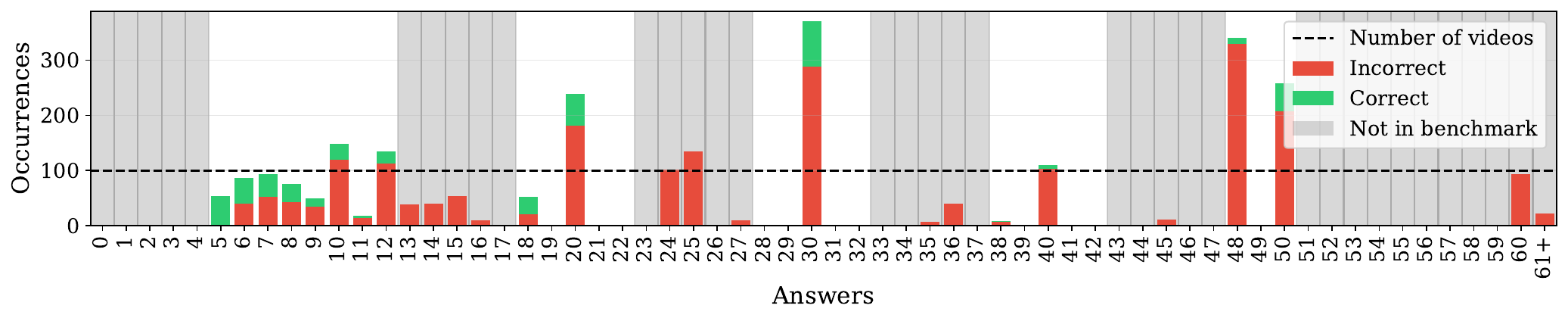} \\
        (k) QWEN3-VL-8B \\[0.15em]

        \includegraphics[width=\textwidth]{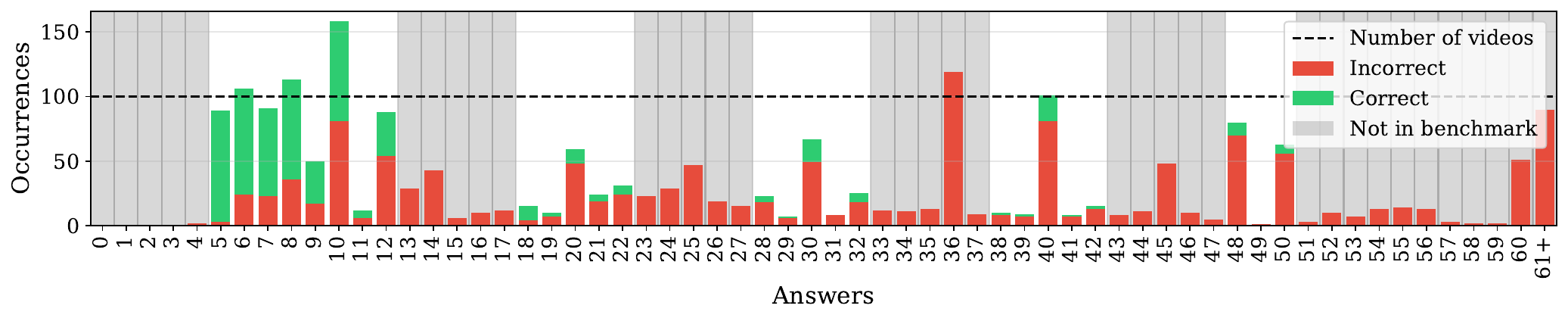} \\
        (l) QWEN3-VL-32B \\[0.15em]
    \end{tabular}

    \caption{Distributions of predicted counts on CountVideoBench. Each bar reports the number of times a given integer value is  predicted by the model across the $2{,}600$ videos.}
    \label{fig:biais_barplots_annexe_6}
\end{figure}

\begin{figure}[H]
    \centering

    \begin{tabular}{c}
        \includegraphics[width=\textwidth]{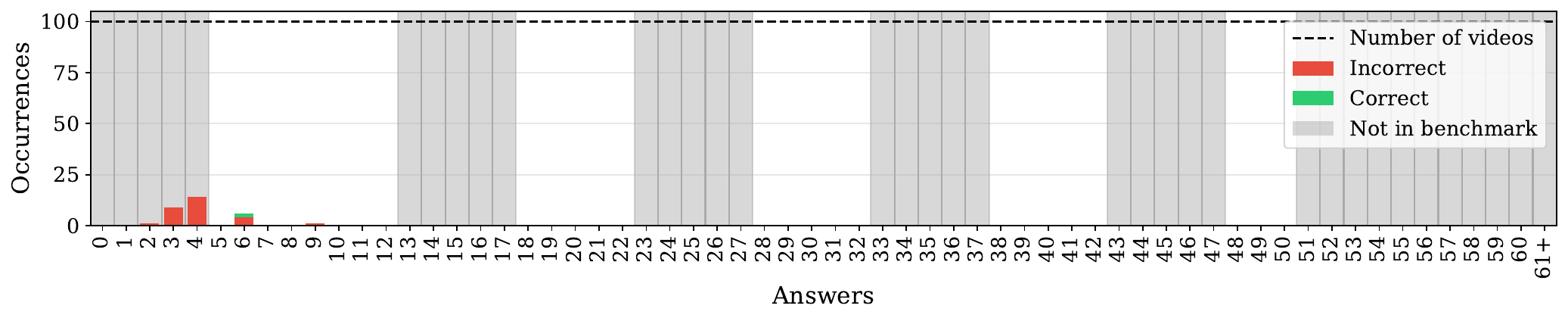} \\
        (m) VideoLLaMA2-7B \\[0.15em]

        \includegraphics[width=\textwidth]{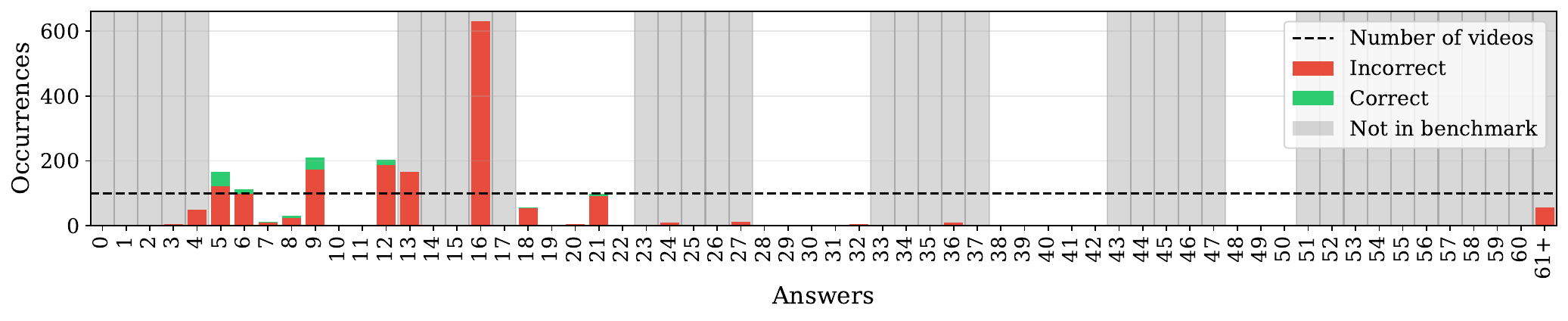} \\
        (n) VideoLLaMA2-72B \\[0.15em]
    \end{tabular}

    \caption{Distributions of predicted counts on CountVideoBench. Each bar reports the number of times a given integer value is  predicted by the model across the $2{,}600$ videos.}
    \label{fig:biais_barplots_annexe_7}
\end{figure}

\begin{figure}[H]
    \centering

    \begin{tabular}{c}
        \includegraphics[width=\textwidth]{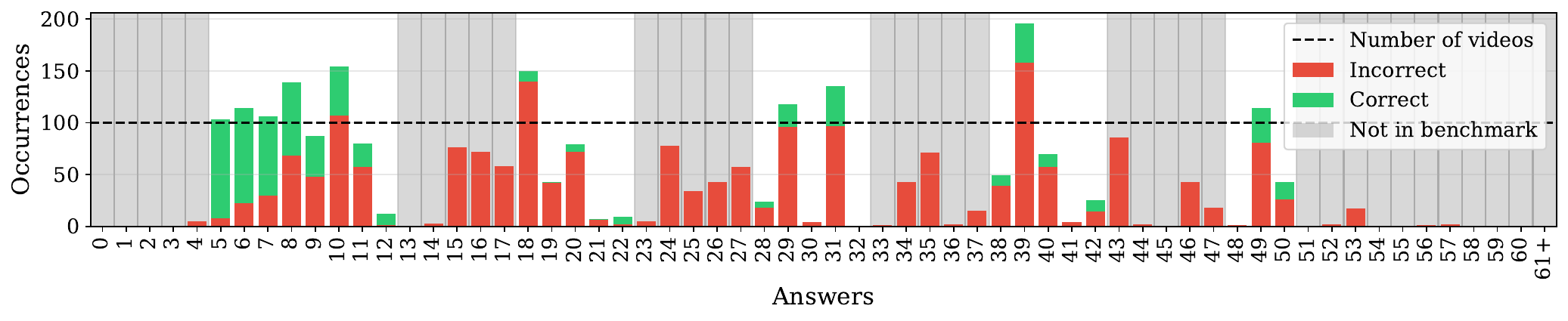} \\
        (o) GPT-5.4 \\[0.15em]%

        \includegraphics[width=\textwidth]{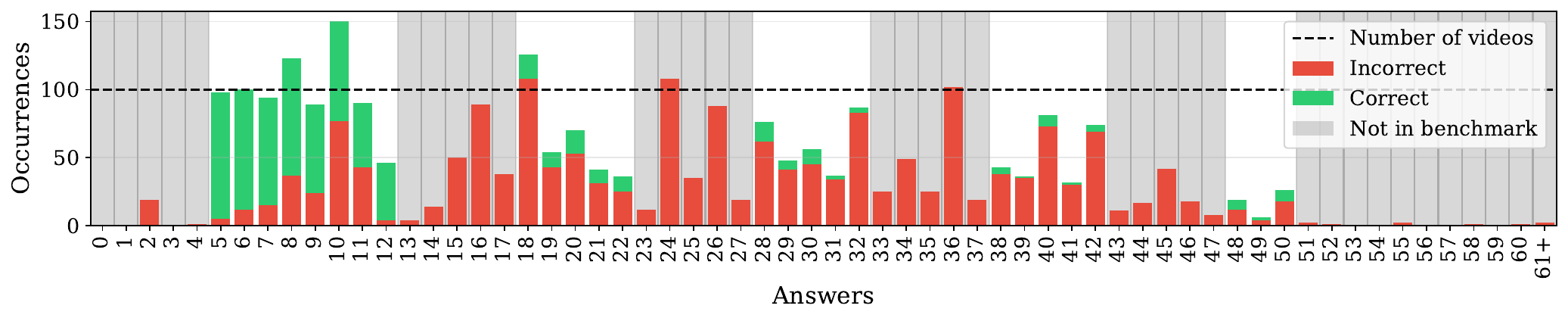} \\
        (p) Gemini-3.1-pro
    \end{tabular}

    \caption{Distributions of predicted counts on CountVideoBench. Each bar reports the number of times a given integer value is  predicted by the model across the $2{,}600$ videos.}
    \label{fig:biais_barplots_annexe_8}
\end{figure}

\section{Prompt Robustness}
\label{app:prompts}

All results in the paper use a single question formulation, fixed in advance.
Since the preferred-value and peak-and-collapse behavior of Section~\ref{sec:bias} is our central finding, we verify that it is not an artefact of that particular wording.

\inlinehead{Prompts.} We evaluated three formulations of the same question:

\begin{itemize}\itemsep0pt
\item[P1] \emph{``How many [color] [shape]s are in this video?''} (used
throughout the paper)
\item[P2] \emph{``Count the [color] [shape]s shown in this video.''}
\item[P3] \emph{``What is the total number of [color] [shape]s in the video?''}
\end{itemize}

Three models were evaluated on all three: InternVL2.5-8B, LLaVA-Video-7B and NVILA-15B. Each of the nine combinations covers the full set of $2{,}600$ videos under identical conditions (same ten sampled frames, same decoding settings, same answer-extraction procedure), so that the prompt is the only varying factor.

\inlinehead{Aggregate performance varies without a common direction.}
Table~\ref{tab:prompts} reports the metrics of Section~\ref{sec:metrics} for the nine combinations. Exact Match spans $10.1$--$19.0\%$ for InternVL2.5-8B, $18.7$--$26.8\%$ for LLaVA-Video-7B and $7.3$--$12.5\%$ for NVILA-15B. No formulation is uniformly preferable: P3 is the best of the three for InternVL2.5-8B and the worst for the other two models. The prompt used in the paper is therefore not one selected for favourable results, and aggregate scores are sensitive to wording in a model-dependent way.

\begin{table}[t]
\centering
\caption{Prompt robustness. All nine combinations are evaluated on the full
$2{,}600$ videos under identical conditions. EM, Acc@10\% and Coverage are in percent; MAE and RMSE in object-count units.}
\label{tab:prompts}
\begin{tabular}{llccccccc}
\toprule
Model & Prompt & EM & Acc@10\% & MAE & RMSE & Cov. & $|\mathcal{S}|$ & $\mathrm{PP}$ \\
\midrule
\multirow{3}{*}{InternVL2.5-8B}
 & P1 & 14.0 & 38.2 & 15.8 & 28.5 &  97.9 & 16 &  9.4 \\
 & P2 & 10.1 & 27.2 &  7.3 & 11.7 & 100.0 & 29 & 16.0 \\
 & P3 & 19.0 & 44.8 &  6.6 & 13.8 &  96.5 & 18 &  9.9 \\
\midrule
\multirow{3}{*}{LLaVA-Video-7B}
 & P1 & 26.8 & 56.3 &  3.7 &  6.1 &  98.9 & 16 &  8.0 \\
 & P2 & 21.7 & 44.3 &  4.3 &  6.5 &  99.0 & 20 & 11.4 \\
 & P3 & 18.7 & 33.6 &  5.1 &  8.0 &  87.4 & 18 & 12.1 \\
\midrule
\multirow{3}{*}{NVILA-15B}
 & P1 & 12.5 & 45.4 &  3.5 &  6.1 & 100.0 & 27 & 14.3 \\
 & P2 &  9.6 & 37.6 &  4.8 &  8.8 & 100.0 & 29 & 15.5 \\
 & P3 &  7.3 & 22.5 &  7.0 & 11.8 & 100.0 & 22 & 10.5 \\
\bottomrule
\end{tabular}
\end{table}

\begin{figure}[t]
\centering
\includegraphics[width=\linewidth]{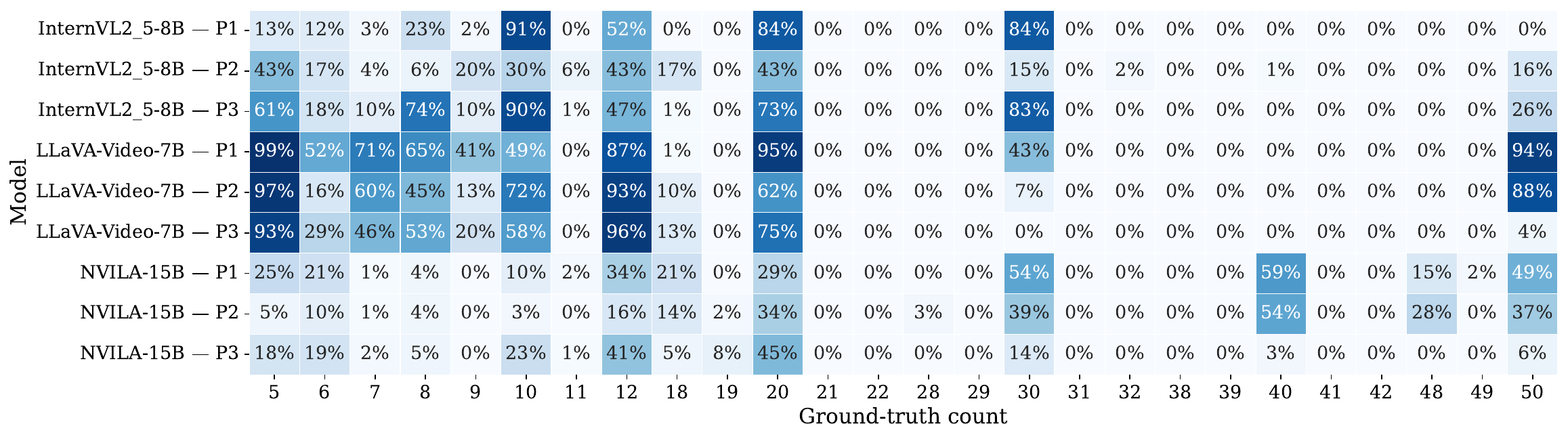}
\caption{Exact Match by ground-truth count for three models under three question formulations (P1--P3, Appendix~\ref{app:prompts}; P1 is the prompt used throughout the paper).
Rows are grouped by model, prompts in order within each group.
Aggregate accuracy differs across rows, but every row alternates between a small number of high-accuracy cardinalities and neighboring values at or near zero.
Eight cardinalities ($21$, $22$, $29$, $31$, $38$, $39$, $41$, $42$) are at exactly $0\%$ in all nine rows.}
\label{fig:prompt_heatmap}
\end{figure}

\inlinehead{The cardinality profile remains structured.}
Figure~\ref{fig:prompt_heatmap} shows Exact Match against ground-truth count for the nine combinations. All of them exhibit the same structure: isolated peaks at a small number of cardinalities, surrounded by collapses at adjacent values.

The invariance is sharpest on the cardinalities that are never solved. Eight values ($21$, $22$, $29$, $31$, $38$, $39$, $41$ and $42$) are at exactly $0\%$ in all nine combinations, and eleven remain below $5\%$. LLaVA-Video-7B, whose inability to produce the answer $40$ is analysed in Section~\ref{sec:bias}, scores $0\%$ at that cardinality under all three formulations. Reformulating the question redistributes accuracy among the cardinalities a model already reaches; it does not make a new one reachable.

\inlinehead{What varies is amplitude.} The largest aggregate drop, NVILA-15B under P3 ($12.5\%$ to $7.3\%$), comes from the flattening of the model's high-cardinality peaks: accuracy at $n = 30$ falls from $54\%$ to $14\%$ and at $n = 40$ from $59\%$ to $3\%$, while the peaks at $10$, $12$ and $20$ rise ($10\%$ to $23\%$, $34\%$ to $41\%$ and $29\%$ to $45\%$). Wording modulates how far up the cardinality range a model's accuracy extends, rather than the alternation of peaks and collapses itself.

\inlinehead{Limitations.} The study covers three formulations on three models under a single decoding configuration. It establishes that the reported structure is not specific to the wording of P1, not that it is invariant to every possible instruction or sampling strategy.

\section{Performance by Object Class}
\label{app:classes}

Section~\ref{sec:design_principles} isolates counting from recognition by restricting each
video to a single object class, drawn from two shapes and five colors. This
appendix reports accuracy broken down by class, to establish what that isolation
does and does not achieve.

\inlinehead{Absolute accuracy depends on the class.} Table~\ref{tab:classes} reports Exact Match per shape and per colour for the sixteen models.
Shape has the larger effect, though it is concentrated on four models: GPT-5.4 reaches $34\%$ on cubes against $17\%$ on spheres, and Qwen3-VL-8B $25\%$ against $13\%$, while NVILA-8B shows the opposite ordering ($3\%$ against $10\%$). Color effects are smaller and less systematic; the widest spreads are InternVL2.5-78B, at $21\%$ on red against $9\%$ on green, and Qwen3-VL-8B, at $27\%$ on yellow against $16\%$ on red and purple.
Twelve of the sixteen models vary by less than five points across colors.

Recognition is therefore reduced but not eliminated by the single-class design:
the ten classes are not equally easy to enumerate, and a model's absolute score
depends in part on which one it is given.

\begin{figure}[t]
    \centering
    \includegraphics[width=0.50\textwidth]{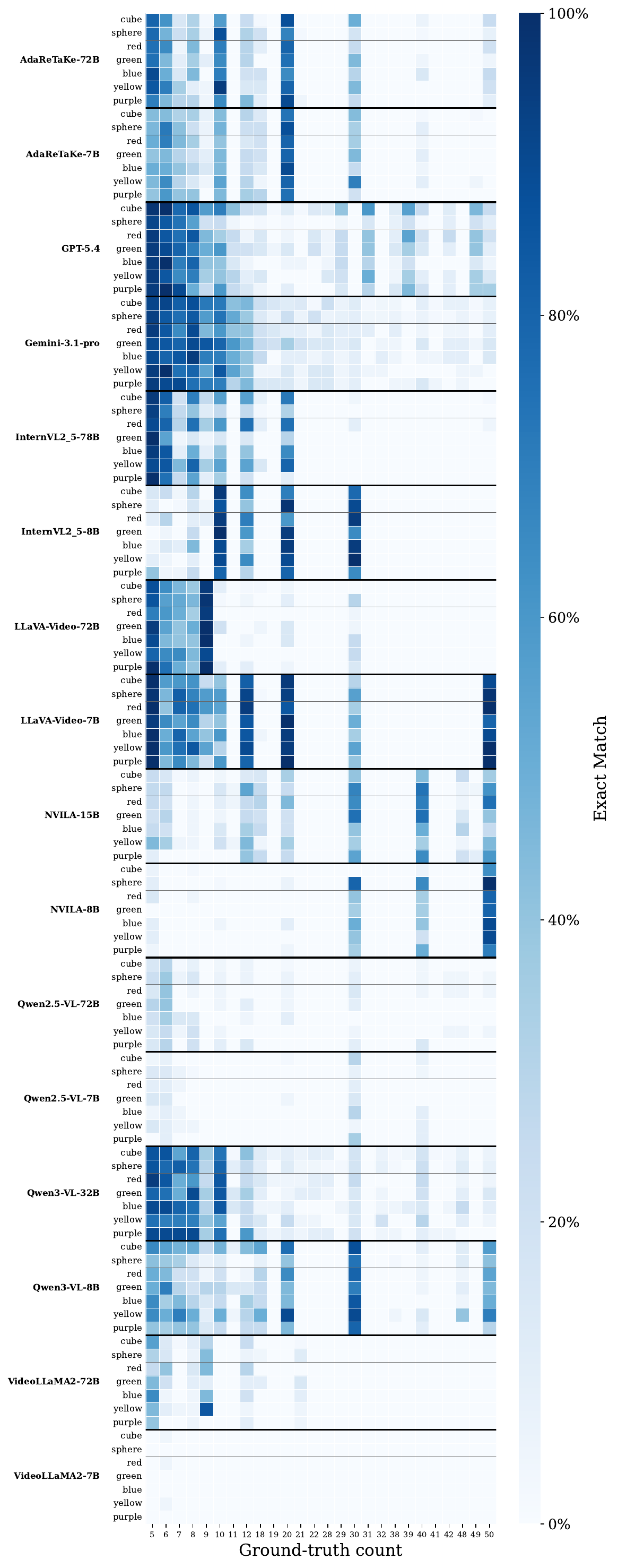}
    \caption{Exact Match against ground-truth count, by object class, for the sixteen models. Each block shows one model, with rows for the two shapes and the five colors; columns are the $26$ cardinalities of the benchmark. Color rows aggregate $20$ videos per cell and shape rows $50$, so differences of five to ten points between rows at a single cardinality are within sampling noise. Within a model the seven rows describe the same profile: the cardinalities a model reaches, and those at which it collapses, do not depend on the class it is asked to count.}
\label{fig:class_heatmap}
\end{figure}

\inlinehead{The cardinality profile does not.}
Figure~\ref{fig:class_heatmap} shows Exact Match against ground-truth count,
separately for each shape and colour, for all sixteen models. Within a model, the
seven profiles describe the same shape. LLaVA-Video-7B is at $0\%$ on $n = 40$ for
every colour and for both shapes, and peaks at $5$, $12$, $20$, $30$ and $50$ in
all seven categories; the peak at $n = 30$ reaches $30\%$ to $56\%$ depending on
the category while $28$, $29$, $31$ and $32$ are at exactly zero in every one of
them. Gemini-3.1-pro declines from high accuracy below $n = 10$ to a
low plateau beyond $n = 28$ in all seven.
Qwen3-VL-32B shows the same
recovery at $n = 40$ ($10\%$ to $30\%$ depending on the category), between two
neighboring cardinalities at or near zero, in six of the seven categories; in
blue the recovery is barely visible ($10\%$, $15\%$ and $0\%$ at $n = 39$, $40$ and $41$).

Each cell of the figure aggregates $20$ videos for the colour rows and $50$ for the
shape rows, so differences of five to ten points between categories at a single
cardinality are within sampling noise and should not be read as class effects.

\inlinehead{Consequence.} Visual class modulates the level of a model's accuracy,
not the structure of its predictions. The preferred-value and peak-and-collapse
behavior characterized in Section~\ref{sec:bias} is present in every category, at
the same cardinalities, for every model whose accuracy is above floor level. The
isolation principle of
Section~\ref{sec:design_principles} is thus sufficient for its purpose: the phenomenon we
report is not an artefact of a particular shape or colour being harder to
recognize.

\begin{table}[t]
\centering
\caption{Exact Match (\%) by object class, over the full $2{,}600$ videos. Shape
columns aggregate $1{,}300$ videos each, colour columns $520$ each.}
\label{tab:classes}
\begin{tabular}{lcc@{\hspace{1.5em}}ccccc}
\toprule
& \multicolumn{2}{c}{Shape} & \multicolumn{5}{c}{Color} \\
\cmidrule(r){2-3}\cmidrule(l){4-8}
Model & Cube & Sphere & Red & Green & Blue & Yellow & Purple \\
\midrule
AdaReTaKe-7B     & 14 & 16 & 15 & 14 & 15 & 16 & 15 \\
AdaReTaKe-72B    & 17 & 16 & 16 & 15 & 17 & 18 & 17 \\
InternVL2.5-8B   & 15 & 13 & 15 & 13 & 15 & 14 & 13 \\
InternVL2.5-78B  & 19 & 13 & 21 &  9 & 15 & 21 & 13 \\
LLaVA-Video-7B   & 25 & 29 & 28 & 26 & 27 & 29 & 25 \\
LLaVA-Video-72B  & 13 & 15 & 12 & 15 & 14 & 14 & 16 \\
NVILA-8B         &  3 & 10 &  7 &  6 &  8 &  6 &  6 \\
NVILA-15B        & 10 & 15 & 15 & 13 & 11 & 12 & 12 \\
Qwen2.5-VL-7B    &  2 &  2 &  1 &  2 &  2 &  2 &  2 \\
Qwen2.5-VL-72B   &  3 &  5 &  4 &  4 &  4 &  3 &  4 \\
Qwen3-VL-8B      & 25 & 13 & 16 & 18 & 18 & 27 & 16 \\
Qwen3-VL-32B     & 22 & 22 & 21 & 22 & 23 & 21 & 24 \\
VideoLLaMA2-7B   &  0 &  0 &  0 &  0 &  0 &  0 &  0 \\
VideoLLaMA2-72B  &  5 &  5 &  6 &  5 &  6 &  6 &  2 \\
GPT-5.4          & 34 & 17 & 28 & 28 & 22 & 25 & 26 \\
Gemini-3.1-pro   & 28 & 27 & 25 & 32 & 28 & 26 & 27 \\
\bottomrule
\end{tabular}
\end{table}

\end{document}